\documentclass[letterpaper, 10 pt, conference]{ieeeconf}  

\IEEEoverridecommandlockouts                              

\usepackage{cite}
\usepackage{graphicx,epstopdf,epsfig}
\DeclareGraphicsExtensions{.eps}
\usepackage{amsmath}
\usepackage{array}
\usepackage{url}
\usepackage{booktabs}
\usepackage{float}
\usepackage[caption=false,font=scriptsize,labelfont=sf,textfont=sf]{subfig}
\graphicspath{{figures/}}
\title{\LARGE \bf
	Synthetic Neural Vision System Design for Motion Pattern Recognition in Dynamic Robot Scenes
}

\author{Qinbing Fu, Cheng Hu, Pengcheng Liu and Shigang Yue
\thanks{This work was supported by the grants of EU Horizon 2020 project STEP2DYNA (691154).}
\thanks{All authors are with the Lincoln Centre for Autonomous Systems (L-CAS), School of Computer Science, University of Lincoln, Lincoln, United Kingdom, LN6 7TS. Email: \{qifu, pliu, chu, syue\}@lincoln.ac.uk}
}

\begin{document}

\maketitle
\thispagestyle{empty}
\pagestyle{empty}

\begin{abstract}
Insects have tiny brains but complicated visual systems for motion perception. A handful of insect visual neurons have been computationally modeled and successfully applied for robotics. How different neurons collaborate on motion perception, is an open question to date. In this paper, we propose a novel embedded vision system in autonomous micro-robots, to recognize motion patterns in dynamic robot scenes. Here, the basic motion patterns are categorized into movements of looming (proximity), recession, translation, and other irrelevant ones. The presented system is a synthetic neural network, which comprises two complementary sub-systems with four spiking neurons -- the lobula giant movement detectors (LGMD1 and LGMD2) in locusts for sensing looming and recession, and the direction selective neurons (DSN-R and DSN-L) in flies for translational motion extraction. Images are transformed to spikes via spatiotemporal computations towards a switch function and decision making mechanisms, in order to invoke proper robot behaviors amongst collision avoidance, tracking and wandering, in dynamic robot scenes. Our robot experiments demonstrated two main contributions: (1) This neural vision system is effective to recognize the basic motion patterns corresponding to timely and proper robot behaviors in dynamic scenes. (2) The arena tests with multi-robots demonstrated the effectiveness in recognizing more abundant motion features for collision detection, which is a great improvement compared with former studies.
\end{abstract}

\section{INTRODUCTION}
\label{introduction}
Building a dynamic vision system in both a robust and efficient manner for motion-sensing in mobile machines, like robots, UAVs and etc, poses a big challenge to modelers. The state-of-the-art computer vision techniques, e.g. \cite{ECCV-2016,multi-motion-2016,CVPR-2016,flownet-2015}, have achieved great improvements on motion/objects detection and tracking. However, these segmentation and/or learning based methods are either computationally costly, or heavily restricted to specific hardware, like event-driven cameras \cite{ECCV-2016}.

In nature, as the results of hundreds of millions of years evolution, animals possess robust visual systems for motion perception. Insects, in particular, have relatively small number of visual neurons, but can navigate smartly through unpredictable and visually cluttered environments. The neural circuits processing visual information in insects are relatively simple compared to those in the human brain, and can be ideal models for optical sensors, as reviewed in \cite{NeuroVisionSensor-2000,FlyingInsects-2010,Serres-2017}. Exploring and modeling of these amazing motion perception neural circuits will significantly advance the applications in vision-based intelligent machines \cite{IROS-LGMDs,Serres-2017}.

\begin{figure}[t]
	\centering
	\includegraphics[width=0.27\textheight]{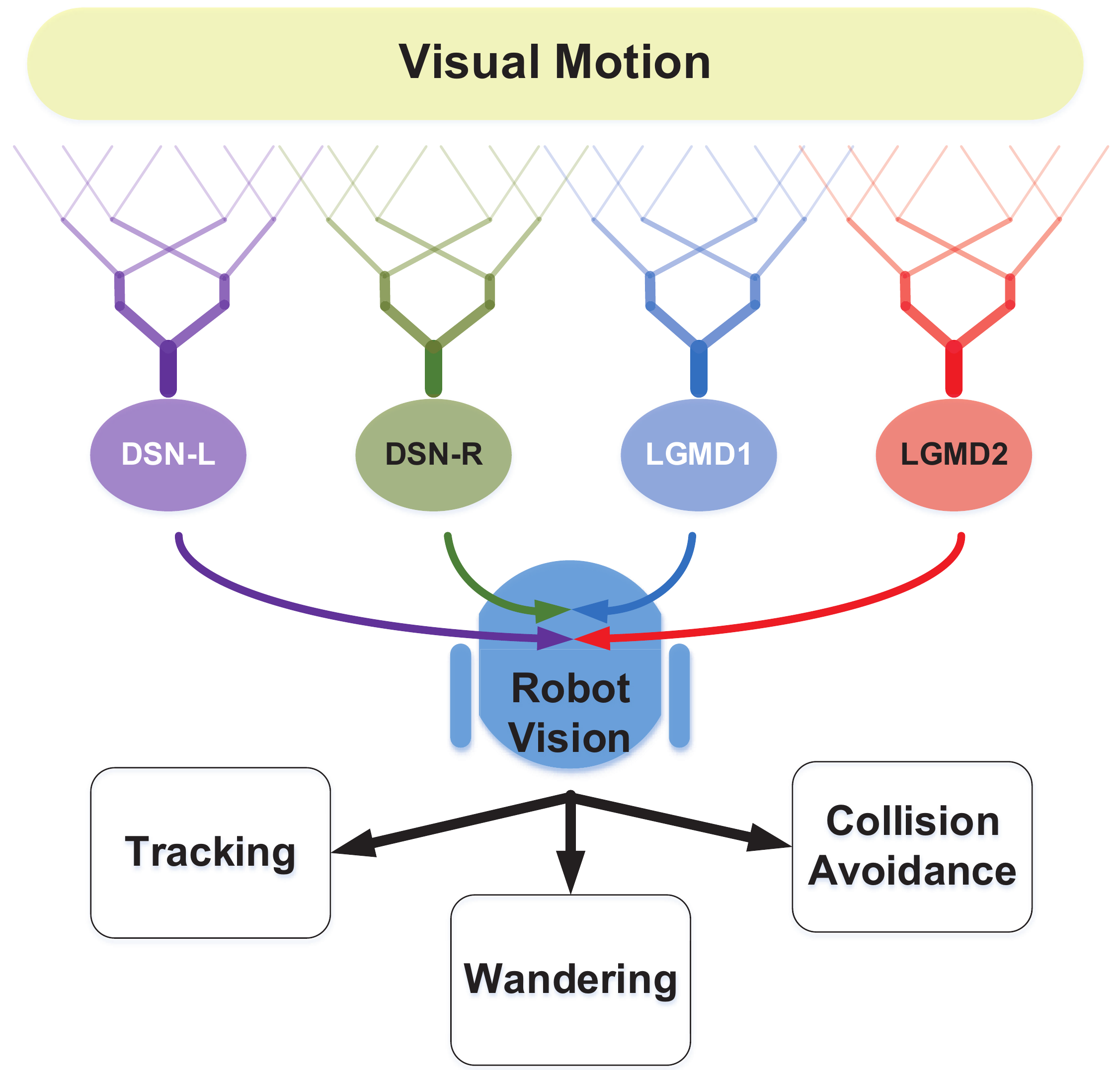}
	\caption{Framework of the proposed biorobotic approach for visual motion features extraction and motion patterns recognition: the inputs to the neural vision system are images captured by a visual modality of the robot; four motion perception neurons (DSN-L, DSN-R, LGMD1, LGMD2) are integrated into the robot vision system to discriminate between different motion cues, in order to invoke distinct behaviors for robot motion control.}
	\label{framework}
\end{figure}

Moreover, on the aspect of visually guided behaviors, insects, like flies, can make correct and timely decisions corresponding to different behaviors, like collision avoidance and target tracking with agile movements in dynamic scenes, while the current mobile robots possess much weaker ability to deal with both motion perception and decision making, especially in dynamic scenes \cite{FlyingInsects-2010,Serres-2017}. In this study, we aim to develop new methods to robotic vision mimicking insects' visual processing strategies, as illustrated in Fig. \ref{framework}.

The lobula giant movement detectors (LGMDs) are large interneurons in the optical lobe of the locust that responds most strongly to fast and direct looming (approaching) objects \cite{LGMDs-2016}. Two LGMDs, i.e. LGMD1 and LGMD2, have been identified by biologists, computationally modeled and successfully applied for collision detection in ground vehicles (e.g. \cite{LGMD-car-2017}), and mobile robots (e.g. \cite{LGMD1-robot2010,Colias-Hu,LGMD2-Fu,LGMD2-BMVC,IROS-LGMDs}). However, through previous biorobotic studies \cite{LGMD2-BMVC,IROS-LGMDs}, we found that the LGMDs-based collision detection models also respond to nearby translating objects. The behavior of collision avoidance is usually triggered by these translational motion patterns, especially in dynamic robot scenes, the situation of which rarely happens in insects \cite{LGMD1-Yue2006}.

To solve this problem, we explored a neuromorphic solution motivated by the direction selective neurons (DSNs) in the fly's visual circuits \cite{DSN-IJCNN}. These visual neurons are only sensitive to wide-field translational motion rather than proximity and recession of objects \cite{Circuit-motion,Borst-common}, which can be ideal neural systems to sense translating objects. The computational visual neural network proposed in \cite{DSN-IJCNN} has demonstrated also the complementary functionality of DSNs to both the LGMD1 and the LGMD2 for motion perception.

Most importantly, via the experience of computationally modeling the LGMDs and the DSNs visual neural networks, we found conspicuous commonality between the model structures of the collision and the translation sensitive neural systems. These bio-plausible models can share some similar signal processing strategies. Recent biological studies have also demonstrated the common circuit design of motion detectors in different animal species \cite{Circuit-motion,Borst-common}. However, these visual neurons each have specific selectivity to different motion features. More specifically, in the locusts, the LGMD1 can respond to the looming of either lighter or darker objects compared to the background, while the LGMD2 is only sensitive to the looming of darker objects \cite{LGMDs-2016}. Such different collision selectivity has been achieved by the modeling of ON and OFF mechanisms \cite{LGMD2-Fu}. With similar ideas, the functionality of the DSNs in the flies, with the direction selectivity to four cardinally directional translations, has been realized by the modeling of ensembles of Reichardt detectors \cite{EMD-1989} in separated ON and OFF pathways \cite{DSN-IJCNN}.

How these different neurons collaborate on motion detection, is thus attractive to modelers to construct a dynamic vision system for recognizing more abundant motion features. There have been a handful of computational studies on incorporating different neural systems. Shigang and Claire developed a model that combines the LGMD1 and the DSNs neural systems, both of which were inspired by the locusts' visual system, to improve the collision detection ability in complex and dynamic driving scenes \cite{LGMD1-Yue2006}. A follow-up study demonstrated the prominent collision-detecting ability of the LGMD1 amongst relevant neural systems \cite{LGMD1-DSN-competing}. Another study demonstrated also the great potential of integrating the locusts' LGMD1 and DSNs \cite{DSN-2013} neural networks for collision detection in driving scenarios, by dividing the field of view into sub-regions processed by different neurons \cite{LGMD-DSNs-Collision}. These works mostly were validated by off-line experiments with video clips as inputs to models. They nevertheless lacked investigation on applications in dynamic robot scenes.

In this research, we apply a biorobotic approach, for the first time integrating visual neuron models inspired by the visual circuits of two insects, to handle visual motion pattern extraction and recognition. Compared with previous works, we will demonstrate the following contributions:
\begin{enumerate}
	\item The proposed biorobotic approach yields simple and effective solutions for fast motion pattern extraction and recognition, which only requires a monocular camera and fewer computational storage capabilities than conventional robotic systems.
	\item The LGMD2 neural system can discriminate darker objects recession from looming well. In the ground robotic scenes, most objects are darker than backgrounds, therefore, the recession pattern can be properly recognized, via combining the LGMD2 model with the LGMD1 model.
	\item The two DSNs neural systems largely enhance the collision selectivity by extracting translational movements in two horizontal directions. Our arena tests demonstrated a great improvement to former two studies for collision detection in dynamic robot scenes.
\end{enumerate}

The rest of this paper is organized as follows: the proposed methodologies will be presented in Section \ref{model}. The micro-robot platform and the neural system setting will be introduced in Section \ref{robot and system}. The robot experiments and results will be illustrated in Section \ref{experiments}. Finally, we conclude this study and give future works in Section \ref{conclusion}.

\section{EMBEDDED VISION SYSTEM}
\label{model}
In general, the proposed embedded vision system consists of two main parts: visual motion extraction and motion pattern recognition. The former comprises four neuron models with low-level spatiotemporal computations in a feed-forward structure. The latter is composed of a switch function and decision making mechanisms, for visually guided robot motion control.

With respect to our former studies on locusts' collision-detecting neurons \cite{LGMD2-Fu,IROS-LGMDs}, and flies' direction selective neurons \cite{DSN-IJCNN,ROBIO-2017}, we highlight the functionality of separated ON and OFF visual pathways, encoding onset and offset response, respectively. Such ON/OFF mechanisms contribute significantly to separate the different selectivity between the LGMD1 and the LGMD2 neural systems, and match well the underlying signal processing circuits in the fly's preliminary visual system.

\subsection{Motion Feature Extraction}
\label{motion-extraction}
The neural system for motion feature extraction is constituted by four computational neuropile layers. All four neurons possess mostly same spatiotemporal computations in the first two computational layers.

\begin{figure*}[t]
	\centering
	\includegraphics[width=0.76\linewidth]{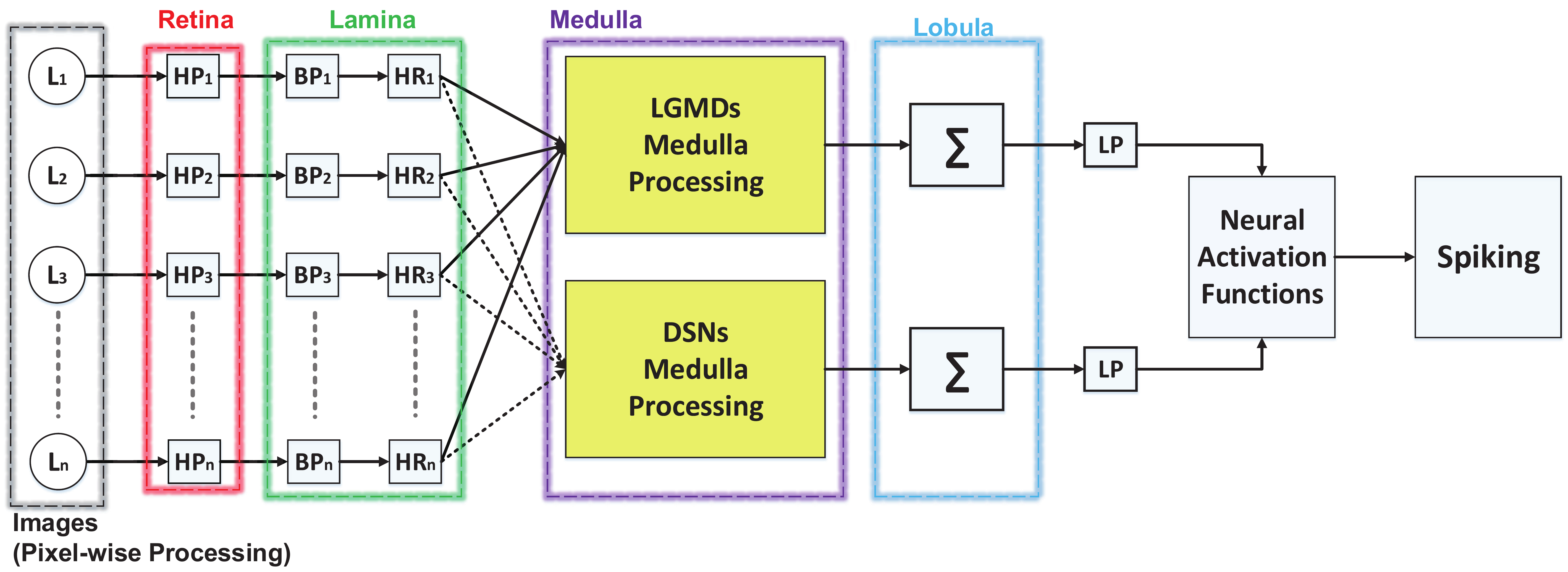}
	\caption{Schematic diagram of the synthetic neural system, with signal processing throughout four computational neuropile layers (retina, lamina, medulla and lobula): the LGMDs and the DSNs models share the same visual processing in the retina and the lamina layers; the different motion feature selectivity is generated in the medulla layer by distinct spatiotemporal computations; the lobula layer integrates local motion, spatially. $L$ is the gray-scale luminance, $HP$, $BP$, $LP$, $HR$ are short for high-pass, band-pass, low-pass filters and half-wave rectifier.}
	\label{visual-model}
\end{figure*}

\paragraph{Computational Retina Layer}
\label{retina}
In the first retina layer, there are photoreceptors ($P$) arranged in a 2D matrix form. As shown in Fig. \ref{visual-model}, the photoreceptors retrieve gray-scale and pixel-wise luminance ($L$), then computes initially motion information by first-order high-pass filters ($HP$), temporally:
\begin{equation}
P(x,y,t) = L(x,y,t) - L(x,y,t-1) + \sum_{i}^{N_i}a_i \cdot P(x,y,t-i)
\label{high-pass}
\end{equation}
where $x$, $y$ are the abscissa and ordinate. $t$ indicates the time sequence in frames. The luminance change could last for a short while of $N_i$ number of frames.  We define a coefficient $a_i$ to be calculated by $a_i = (1 + e^{u \cdot i})^{-1}$ and $u = 1$, for the fast decay of residual information in animals' visual circuits.

\paragraph{Computational Lamina Layer}
\label{lamina}
In the second lamina layer, we apply band-pass filters to achieve the edge selectivity to motion features, as well as removing redundant environmental noise, spatially. Two linearly distributed Gauss kernels are used to convolve visual signals, so as to save computational power in case of limited resources:
\begin{equation}
P_{e/i}(x,y,t) = P(x,y,t) \overset{x,y}{*} W_{e/i}(x,y)
\label{band-pass-convolve}
\end{equation}
where $\overset{x,y}{*}$ indicates the convolution at each local cell $(x,y)$. The weightings of the excitatory and the inhibitory kernels $W_e$, $W_i$ are given in Table \ref{params}. In addition, the outer inhibitory kernel is with twice size of the inner excitatory kernel. The excitation is subtracted from the inhibition:
\begin{equation}
P^{'}(x,y,t) = P_{e}(x,y,t) - P_{i}(x,y,t)
\label{band-pass-subtract}
\end{equation}
After that, there are ON and OFF polarity interneurons splitting visual information into parallel ON and OFF channels, encoding onset and offset responses, respectively, by the mechanism of half-wave rectifier:
\begin{equation}
\begin{aligned}
&P_{on}(x,y,t) = (P^{'}(x,y,t) + |P^{'}(x,y,t)|) / 2,\\
&P_{off}(x,y,t) = |(P{'}(x,y,t) - |P^{'}(x,y,t)|)| / 2
\end{aligned}
\label{half-wave}
\end{equation}

In this biorobotic study, we also adopt a bio-plausible mechanism to realize an `adaptation state', with a fast onset and slow decay characteristic, which significantly reduces noise in time. Let $X$, $Y$ be short for $P_{on/off}(x,y)$ and delayed signal $D_{on/off}(x,y)$, the mathematic expression of the temporal mechanism is as follows:
\begin{equation}
d Y(t)/d t = \left\{
\begin{aligned}
&(X(t) - Y(t))/\tau_{1},\ \text{if}\ d X(t)/d t \ge 0\\
&(X(t) - Y(t))/\tau_{2},\ \text{if}\ d X(t)/d t < 0,
\end{aligned}
\right.
\label{fdsr-lowpass}
\end{equation}
where $\tau_{1}$ and $\tau_{2}$ are time constants in milliseconds and $\tau_{1} < \tau_{2}$. Then, the filtered signal is subtracted to the original one:
\begin{equation}
\begin{aligned}
&F_{on}(x,y,t) = P_{on}(x,y,t) - D_{on}(x,y,t),\\
&F_{off}(x,y,t) = P_{off}(x,y,t) - D_{off}(x,y,t)
\end{aligned}
\label{fdsr-subtraction}
\end{equation}

\paragraph{Computational Medulla Layer}
\label{medulla}
The third computational medulla layer is of great importance in generating different motion feature selectivity. Concretely, the LGMD1 and the LGMD2 neurons are directionally selective to movements in depth, i.e. looming and recession, while the DSN-R and the DSN-L neurons are directionally selective to movements in two horizontal directions, i.e. rightward and leftward translations, respectively. Intuitively, the functionality of the DSNs provides perfect complement to the functionality of the LGMDs. Moreover, compared with former modeling studies \cite{LGMD-DSNs-Collision,LGMD1-DSN-competing,LGMD1-Yue2006}, the specific looming selectivity of the LGMD2 neuron to darker objects only, could advance the discrimination between looming and recession movements.
\begin{figure*}[t]
	\centering
	\includegraphics[width=0.78\linewidth]{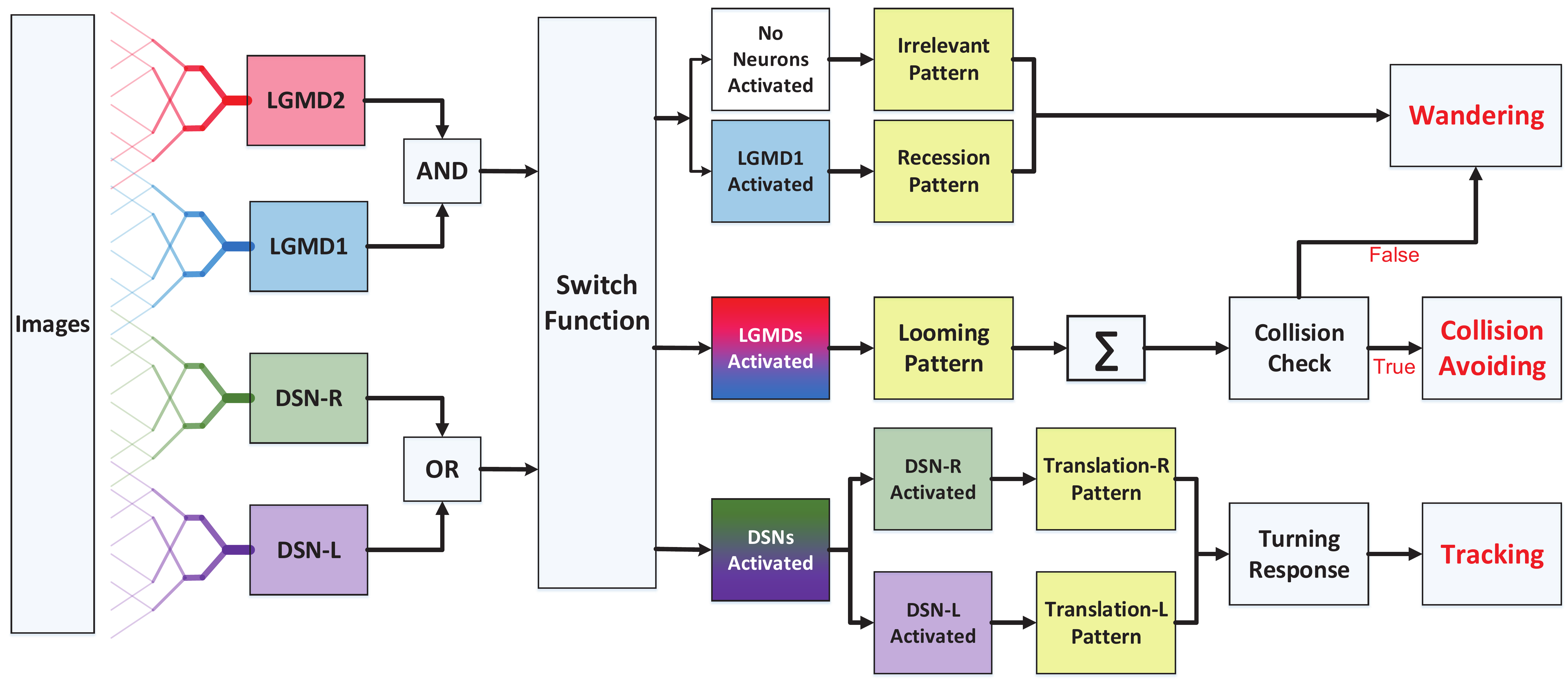}
	\caption{Diagram of the motion pattern recognition strategy, including motion feature extraction by four neurons, recognition and decision making mechanisms, as well as simple robot behaviors. \textbf{The synthetic neural vision system processes visual information and conducts robot motion frame-by-frame in a feed-forward structure.}}
	\label{whole-system}
\end{figure*}

First, for the modeling of LGMDs, both neurons detect potential collision by reacting to expanding edges of objects. In the ON pathway, the inhibition is formed by convolving surrounding delayed excitations, while in the OFF pathway, the excitation is formed by convolving surrounding delayed inhibitions \cite{IROS-LGMDs}. There are local summation cells integrating the local excitations and inhibitions from the dual-pathways:
\begin{equation}
\begin{aligned}
&S_{on}(x,y,t) = F_{on}(x,y,t) - w_1 \cdot D^{'}_{on}(x,y,t) \overset{x,y}{*} W_l(x,y),\\
&S_{off}(x,y,t) = D^{'}_{off}(x,y,t) \overset{x,y}{*} W_l(x,y) - w_2 \cdot F_{off}(x,y,t)
\end{aligned}
\label{lgmds-ei-s}
\end{equation}
where $W_l$ is a convolution kernel. $w_1$, $w_2$ are two local bias to suppress inhibitory flows. $D^{'}_{on/off}$ is delayed by $F_{on/off}$ similarly to Eq. \ref{fdsr-lowpass}, yet with a dynamic time parameter $\tau_s$ in milliseconds. Importantly, the following interactions between ON and OFF summation cells realize the different looming selectivity between LGMD1 and LGMD2 neurons:
\begin{equation}
S = \theta_1 \cdot S_{on} + \theta_2 \cdot S_{off} + \theta_3 \cdot S_{on} \cdot S_{off}
\label{supralinear}
\end{equation}
where $\{\theta_1, \theta_2, \theta_3\}$ indicates the combination of term coefficients, in order to mediate the excitations from either ON/OFF pathways. In case of LGMD2, the excitations from the ON channels are rigorously suppressed, forming the looming selectivity to dark objects only \cite{IROS-LGMDs}.

Second, on the aspect of modeling the DSNs, we design ensembles of ON/OFF local motion detectors, each combination of which is composed of a pairwise Reichardt detectors:
\begin{equation}
\begin{aligned}
ON(x,y,t) &= \sum_{i=d}^{d \cdot N_{c}}(D'_{on}(x,y,t) \cdot F_{on}(x+i,y,t)\\ &- D'_{on}(x+i,y,t) \cdot F_{on}(x,y,t)),\\
OFF(x,y,t) &= \sum_{i=d}^{d \cdot N_{c}}(D'_{off}(x,y,t) \cdot F_{off}(x+i,y,t)\\ &- D'_{off}(x+i,y,t) \cdot F_{off}(x,y,t))
\end{aligned}
\label{dsns-ei-s}
\end{equation}
where $d$ and $N_c$ are the sampling distance between each pairwise detectors and the number of connected ON/OFF cells, respectively. As the robot can only move on a 2D surface, we only calculate the directional motion in two horizontal directions. Compared to the LGMDs model, the spatiotemporal computations in Eq. \ref{dsns-ei-s} can realize the selectivity to translations versus looming and recession features.

\paragraph{Computational Lobula Layer}
\label{lobula}
In the lobula layer, both the DSNs and the LGMDs neuron models integrate all local motion signals from the ON and OFF visual pathways, linearly and spatially \cite{DSN-IJCNN,IROS-LGMDs}. After that, the global excitations are transformed to the membrane potential via sigmoid functions $f(x) = (1 + e^{-|x|/(n \cdot K_{sig})})^{-1}-\Delta_{C}$, as the neural activation functions in Fig. \ref{visual-model}, where $n$ denotes the total number of pixels in the field of view and $K_{sig}, \Delta_{C}$ are scale parameters. The outputs of the LGMD1 and the LGMD2 neurons are both normalized within $[0.5,1)$, whilst the outputs of the DSN-R and the DSN-L neurons are normalized within $[0,1)$ and $(-1,0]$, respectively.

\paragraph{Spiking Mechanism}
\label{spiking}
In this biorobotic approach, we implement these four visual neurons as spiking neurons. The membrane potential is transformed to spikes at each frame, exponentially:
\begin{equation}
S^{spike}(t) = \left \lfloor{e^{[K_{sp} \cdot (U(t) - T_{sp})]}}\right \rfloor
\label{spiking mechanism}
\end{equation}
where $\left \lfloor{x}\right \rfloor$ indicates a `floor' function to return the largest integer less than or equal to the specified input $x$. $K_{sp}$ and $T_{sp}$ denote a coefficient and the spiking threshold. $U(t)$ is the membrane potential of either neurons. As a result, more than one spikes could be generated at each frame.

\subsection{Motion Pattern Recognition}
\label{motion-recognition}
Generally speaking, we highlight a neural competition between the LGMDs and the DSNs in the motion pattern recognition mechanisms. The activation of either DSN-R or DSN-L neurons will rigorously inhibit both the LGMD1 and the LGMD2 neurons, and vice versa. As shown in Fig. \ref{whole-system}, the generated spikes of these four neurons are conveyed to logical operations and a switch function, which generates three outcomes:
\begin{enumerate}
	\item The situations of no neurons activated or LGMD1 neuron activated only, correspond to an `irrelevant motion pattern' or a potential `recession pattern', respectively, followed by a `wandering' state for robot motion.
	\item Once the LGMDs win the competition with higher spiking rate, a potential `looming pattern' is given. However, a confirmation of collision detection should meet the following requirement:
	\begin{equation}
	Col(t) = \left\{
	\begin{aligned}
	&\text{true},\ \text{if}\ \sum_{i=t-N_{t}}^{t}S^{spike}(i) \ge N_{sp}\\
	&\text{false},\ \text{otherwise}
	\end{aligned}
	\right.
	\label{collision recognition}
	\end{equation}
	where $N_{sp}$, $N_{t}$ denote the number of successive spikes and frames. If the collision is verified, an `avoidance' behavior will be triggered; otherwise, the robot will remain wandering.
	\item If the DSNs represent higher spiking frequency, either a `rightward translation' or a `leftward translation' pattern is recognized, corresponding to a `turning response' computed as follows:
	\begin{equation}
	\begin{aligned}
	&TR(t) = \sigma_1 \cdot U^{dsn}(t),\\
	&\text{then},\ d \{TR'(t)\} / d t = (TR(t) - TR'(t)) / \tau_{3}
	\end{aligned}
	\label{turning-response}
	\end{equation}
	where $\sigma_1$ is a term coefficient and $\tau_{3}$ is a time constant in the low-pass filtering. As a result, a `tracking' behavior will be triggered.
\end{enumerate}

\subsection{Robot Motion Control}
\label{motion-control}
On the aspect of motion control strategies, a robot agent is given an initial speed $v_i$. If the current state is either the `wandering' or the `tracking', the motor powers of the right ($P_R$) and left ($P_L$) wheels can be described as follows:
\begin{equation}
\begin{aligned}
&P_R(t) = g_v \cdot v_i(t) - g_w \cdot TR'(t),\\
&P_L(t) = g_v \cdot v_i(t) + g_w \cdot TR'(t)
\end{aligned}
\label{motor-power}
\end{equation}
$g_v$ and $g_w$ are gain values that control motion efficiency. Otherwise, if the state is the `collision avoiding', we implement a motion sequence in the robot agent to turn around with a radian over $\pi$, randomly to the left or right.

\section{ROBOT AND SYSTEM CONFIGURATION}
\label{robot and system}
\begin{table}[t]
	\caption{The Predefined parameters}
	\centering
	\begin{tabular}{|ll|ll|ll|}
		\toprule
		Name          &Value     &Name       &Value      &Name   &Value\\
		\hline
		$N_c,d$         &$2\sim4$	 &$K_{sp}$		 &$1\sim6$   &$W_{e/i}$   &$1/(4\sim128)$\\
		$W_l$    	  &$1/(4\sim8)$      &$w_1$     &$0.3$       &$N_i$      &$2$\\
		$\tau_{1}$	  &$1$     	&$\tau_{2}$ &$100$ms    &$K_{sig}$ &$0.1\sim0.6$\\
		$\Delta_{C}$  &$0\sim1$     &$w_2$   &$0.6$      &$\sigma_1$ &$15$\\
		$\theta_1,\theta_2$	  &$0\sim1$	  &$\theta_3$ &$0$        &$\tau_{3}$ &$10$\\
		$\tau_s$	  &$10\sim200$	&$g_v$   &$1$        &$g_w$      &$10$\\
		$T_{sp}$	&$0.2,0.7$	&$n$	&$99\times72$	&$N_{sp},N_t$	&$6,4$\\
		\bottomrule
	\end{tabular}
	\label{params}
\end{table}
In this section, we propose the parameters setting of the embedded vision system, and briefly introduce the robot platform. First, the parameters were all decided empirically according to the implementation and optimization on the micro-robot platform. Table \ref{params} lists the set-ups of parameters presented in Section \ref{model}. More specifically, the spiking thresholds for the LGMDs and the DSNs neural systems are $0.7$ and $0.2$ respectively. Importantly, these spiking thresholds and scale parameters $K_{sig}$ in the neural activation functions, $K_{sp}$ in the spiking mechanism greatly affect the spike frequency of these visual neuron. Some parameters could also vary within specific ranges. In this biorobotic approach, we do not apply any learning methods or feedback control.

\begin{figure}[t]
	\centering
	\subfloat[]{\frame{\includegraphics[width=0.2\textwidth]{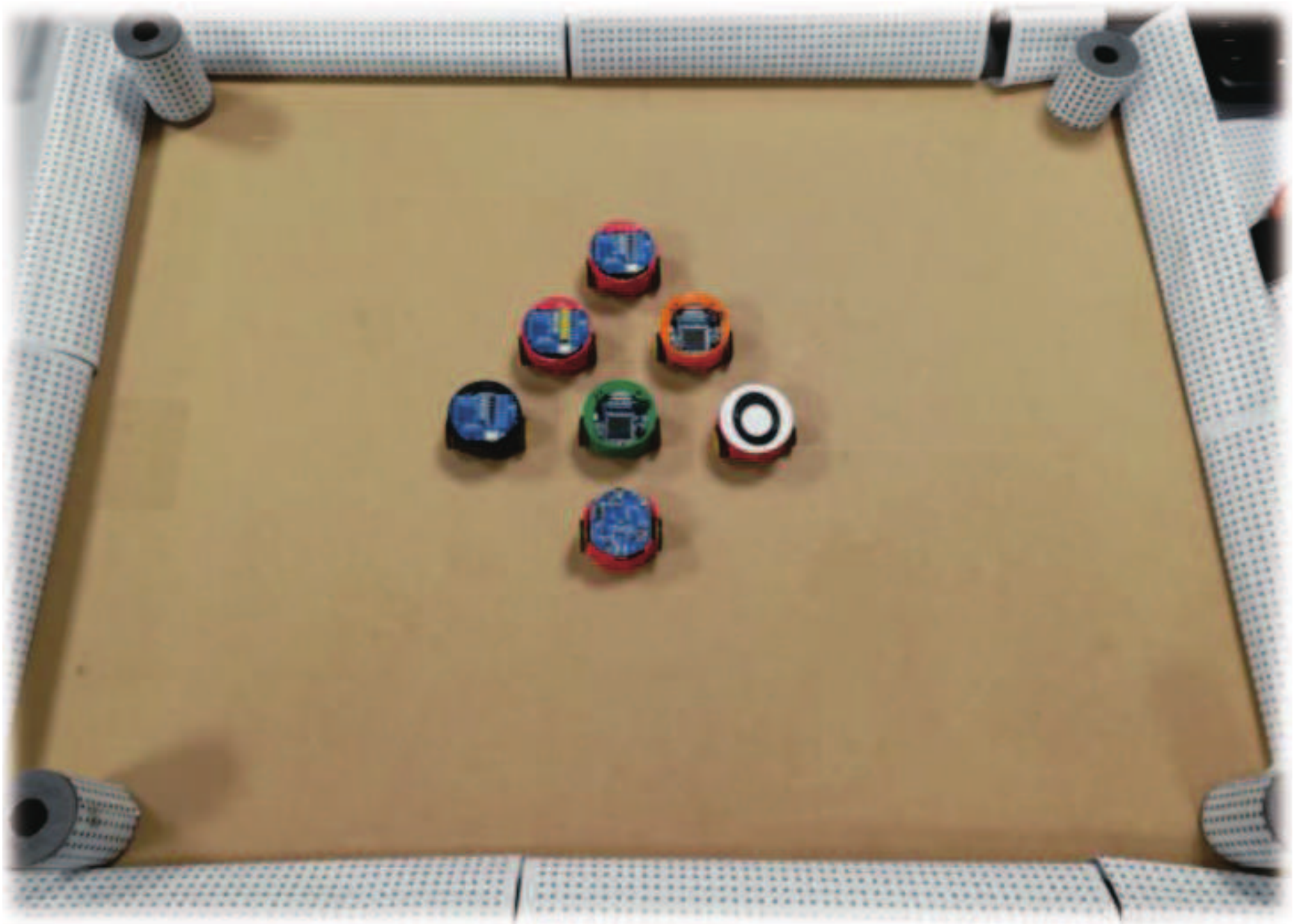}}
		\label{small-arena}}
	\hfil
	\subfloat[]{\frame{\includegraphics[width=0.27\textwidth]{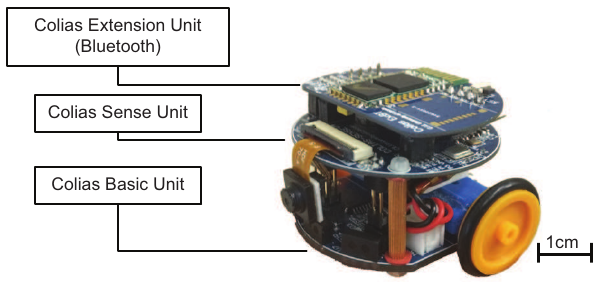}}
		\label{colias-units}}
	\caption{Illustrations of the small arena and the vision-based micro-robot.}
	\label{colias and arena}
\end{figure}

The monocular vision based micro-robot is a low-cost and autonomous ground mobile platform named `\textit{Colias}' \cite{Colias-Hu,IROS-LGMDs}. As illustrated in Fig. \ref{colias-units}, the \textit{Colias} robot has a small footprint of $4$cm in diameter and $3$cm in height. Two DC-motors are driven differentially and provide the platform with a maximum speed of roughly $35$cm/s. A $3.7$V, $320$mAh Lithium battery supports the autonomy for $1\sim2$ hours. The \textit{Colias} robot has two main boards: the bottom board includes wheels and battery, working as a motion actuator on 2D surfaces; the upper board supports in-chip image processing with an $OV7670$ camera. Its main processor is an ARM- Cortex M4 based MCU STM32F427, which runs at $180$MHz, with $256K$byte SRAM, $2$Mbyte in-chip Flash. The acquired image is set to $99\times72$ in YUV422 format at $30$ fps. In addition, the field of view can reach approximately $70$ degrees. The only sensor used in this research is the monocular camera. We also used a Bluetooth device, which is connected with the upper board, to retrieve real-time data from the robot. The frame rate of the embedded vision system is between $25\sim 35$Hz, which well fits the requirement of most real-time visual tasks.

We also built a small arena for conducting dynamic robot scenes in arena tests. As illustrated in Fig. \ref{small-arena}, the arena is with $70\times55 cm^2$ in acreage. The peripheries of the arena were decorated with specific patterns, as textures for visual motion perception.

\section{EXPERIMENTS AND RESULTS}
\label{experiments}
In this section, we present our robot experiments and analyze the results. There are mainly two categories of tests: in the open-loop tests, we will firstly demonstrate the neural response of different visual neurons to the four basic motion patterns, as illustrated in Fig. \ref{open-robot-tests}. To verify the effects of four neurons on different motion pattern extraction and recognition, we will demonstrate also the statistical investigations on activations (spiking rates) of these neurons, which are challenged by the four kinds of robot movements, at different constant speeds, repeatedly and respectively (Fig. \ref{open-robot-statistics}). Moreover, we will investigate the influence of angular approaching movements on motion pattern recognition, which are also frequent visual challenges to robots in dynamic scenes (Fig. \ref{angular-approach-tests}).

In the second type of tests, we will demonstrate our arena tests, with multiple \textit{Colias} robots, forming the dynamic robot scenes in the small arena. To highlight the achievements of this biorobotic approach, we will compare the success rate of collision detection to two former studies \cite{IROS-LGMDs,LGMD2-BMVC}, with new motion patterns, that is, translations been identified as non-collision events under identical robot densities. Some video snapshots of the arena tests, captured by a top-down camera\footnote{Available in the attached video demo.}, are shown in Fig. \ref{arena captures}. The success rates of different events in arena tests are given in Table \ref{sr-table}.

\subsection{Open-loop Robot Tests}
\label{open-loop}
\begin{figure*}[t]
	\centering
	\subfloat[Robot Looming]{\frame{\includegraphics[width=0.16\linewidth]{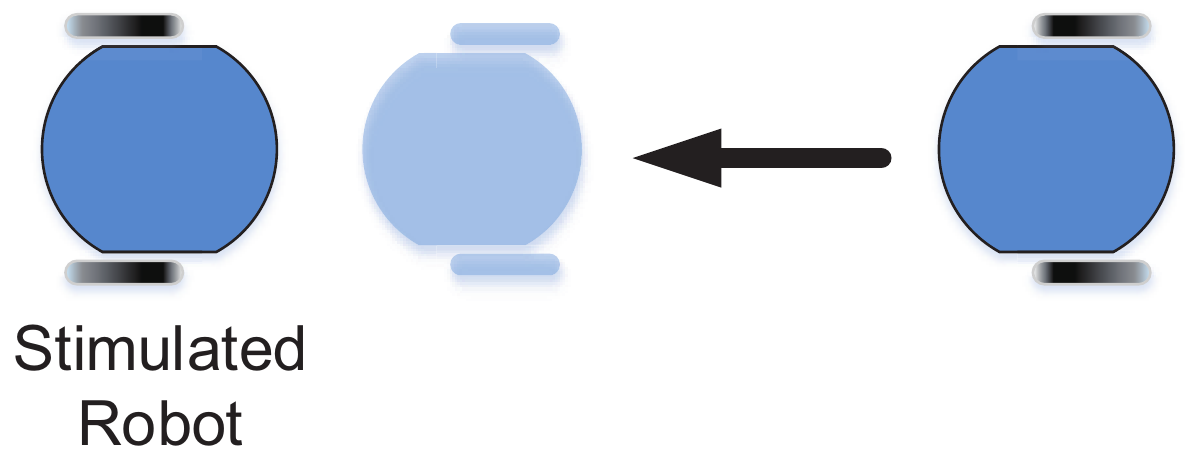}}
		\label{looming-pattern}}
	\hfil
	\hspace{0.1in}
	\subfloat[Robot Recession]{\frame{\includegraphics[width=0.15\linewidth]{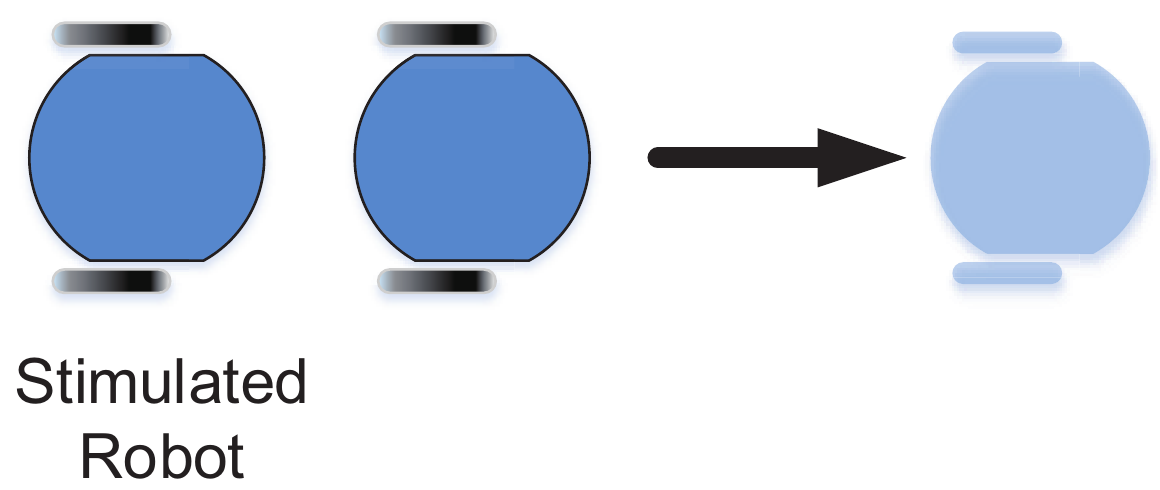}}
		\label{recession-pattern}}
	\hfil
	\hspace{0.1in}
	\subfloat[Robot Trans-R]{\frame{\includegraphics[width=0.11\linewidth]{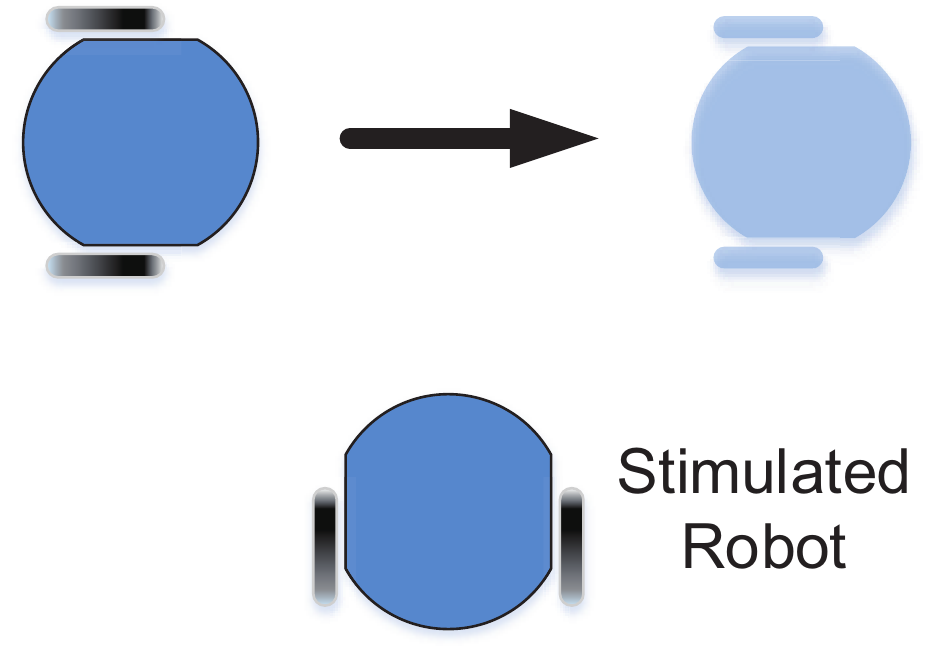}}
		\label{transfb-pattern}}
	\hfil
	\hspace{0.3in}
	\subfloat[Robot Trans-L]{\frame{\includegraphics[width=0.11\linewidth]{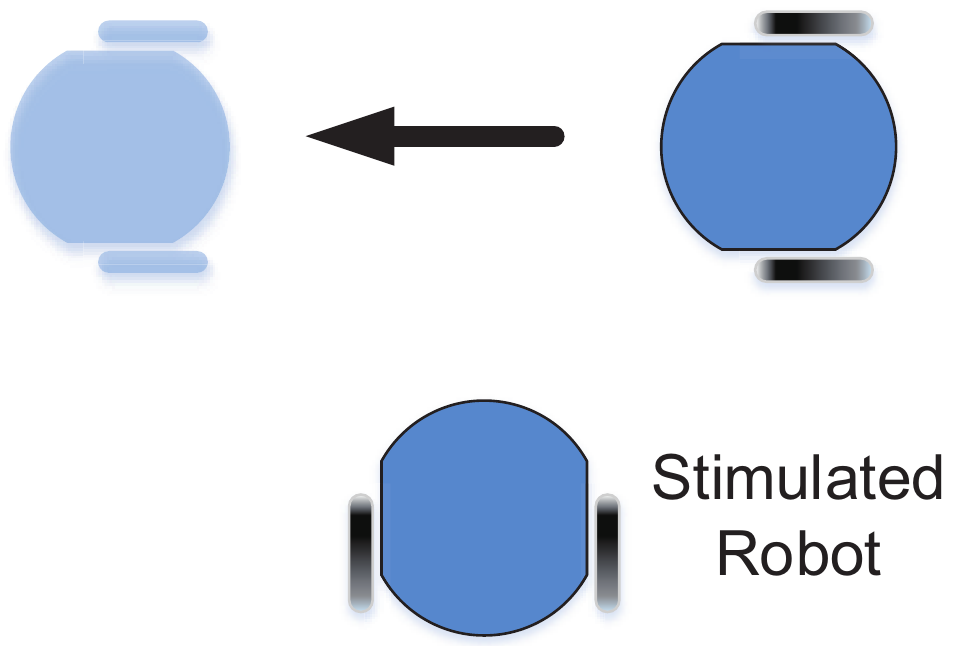}}
		\label{transbf-pattern}}
	\vfill
	\vspace{-0.1in}
	\subfloat{\includegraphics[width=0.24\linewidth]{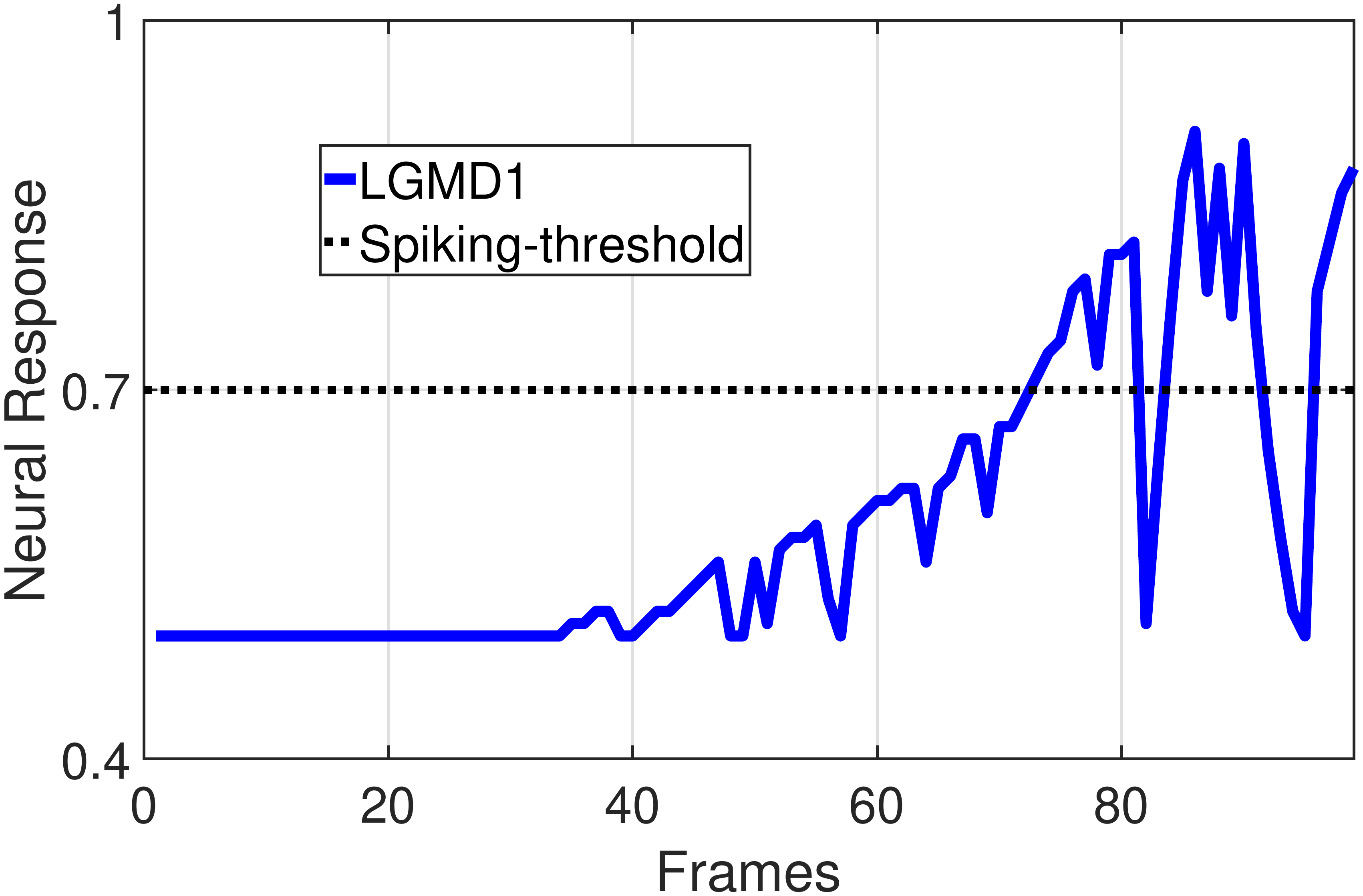}
		\label{looming-lgmd1-potential}}
	\hfil
	\subfloat{\includegraphics[width=0.24\linewidth]{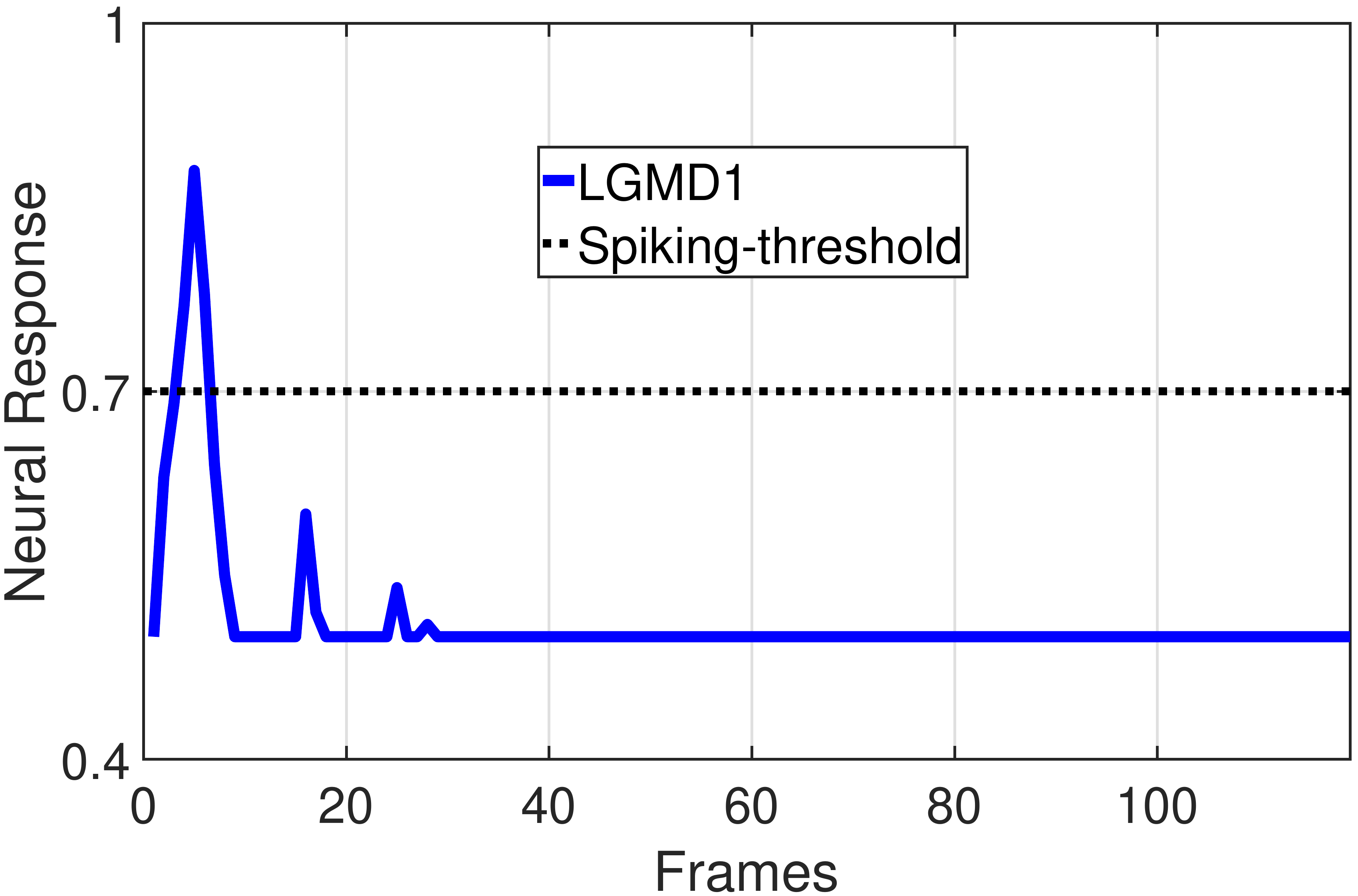}
		\label{recession-lgmd1-potential}}
	\hfil
	\subfloat{\includegraphics[width=0.24\linewidth]{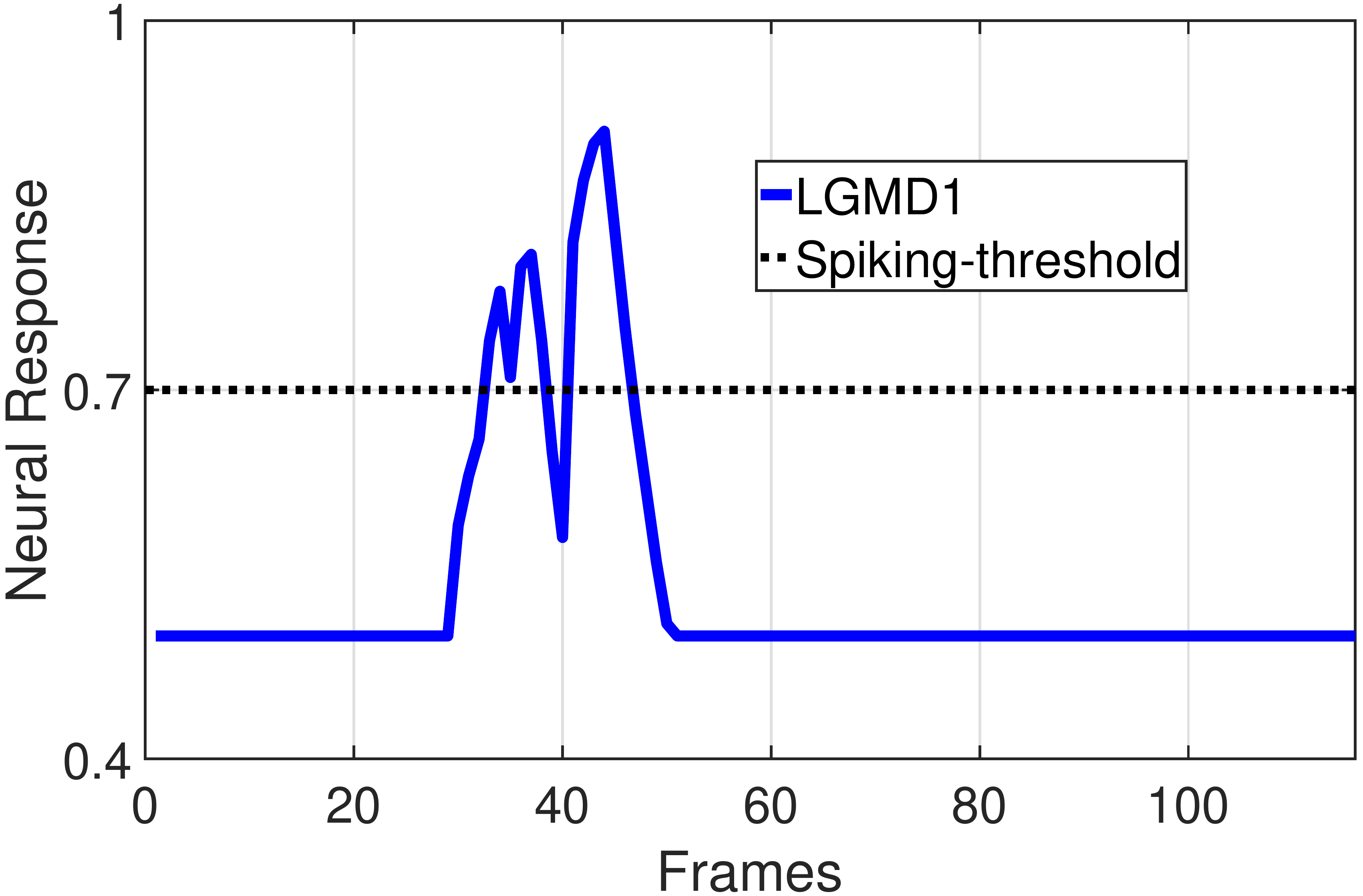}
		\label{translation-fb-lgmd1-potential}}
	\hfil
	\subfloat{\includegraphics[width=0.24\linewidth]{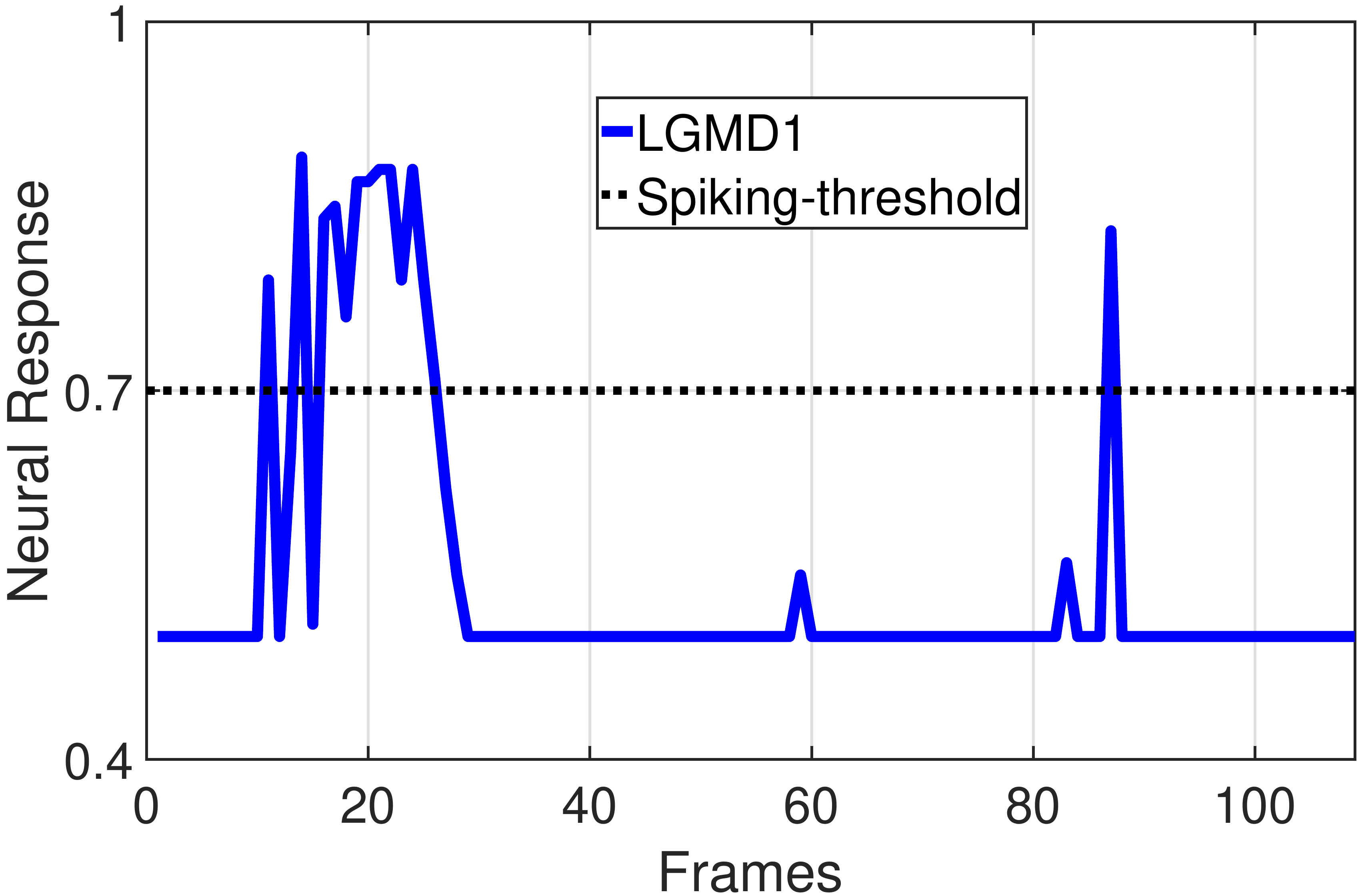}
		\label{translation-bf-lgmd1-potential}}
	\vfill
	\vspace{-0.1in}
	\subfloat{\includegraphics[width=0.24\linewidth]{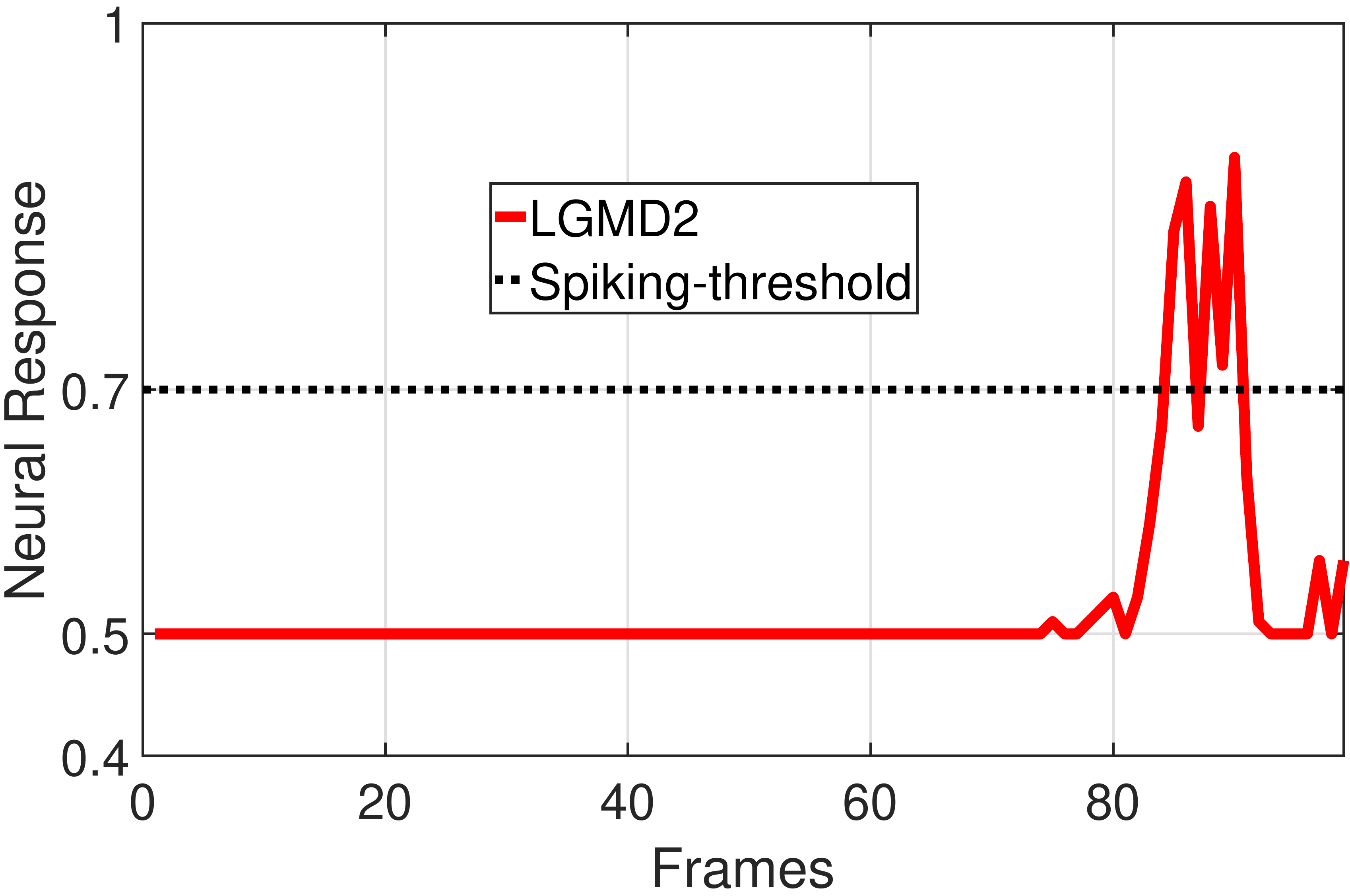}
		\label{looming-lgmd2-potential}}
	\hfil
	\subfloat{\includegraphics[width=0.24\linewidth]{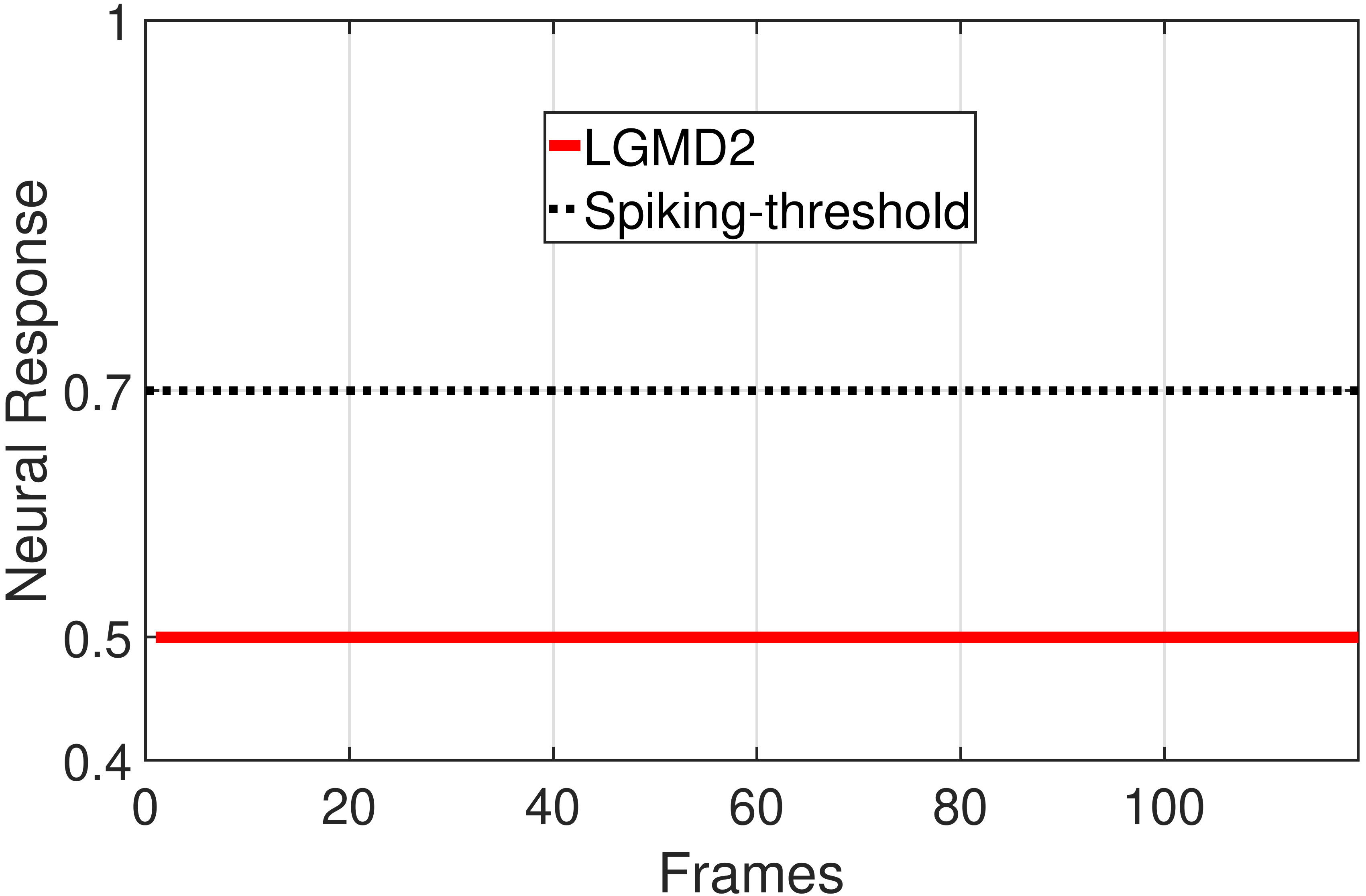}
		\label{recession-lgmd2-potential}}
	\hfil
	\subfloat{\includegraphics[width=0.24\linewidth]{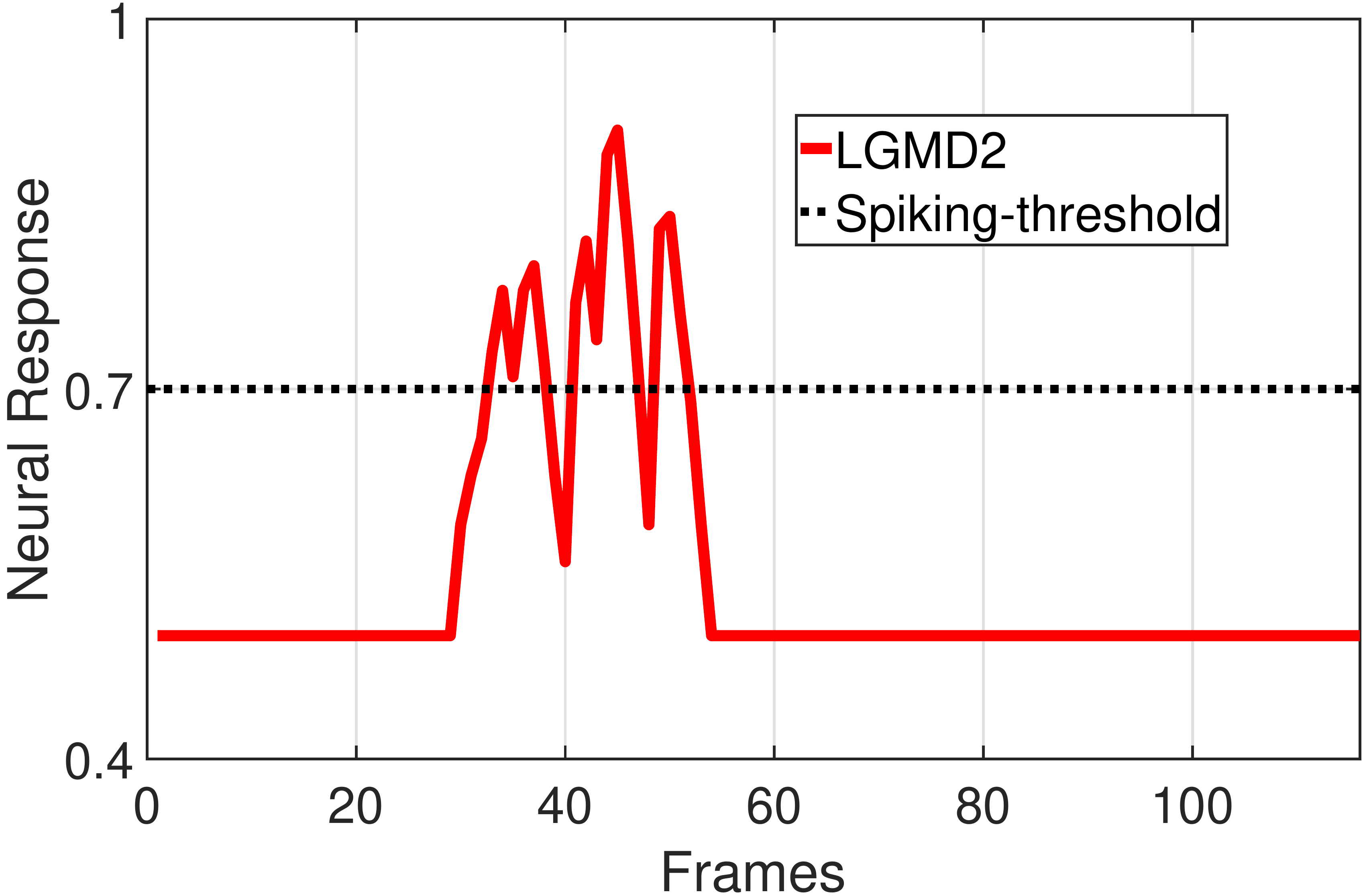}
		\label{translation-fb-lgmd2-potential}}
	\hfil
	\subfloat{\includegraphics[width=0.24\linewidth]{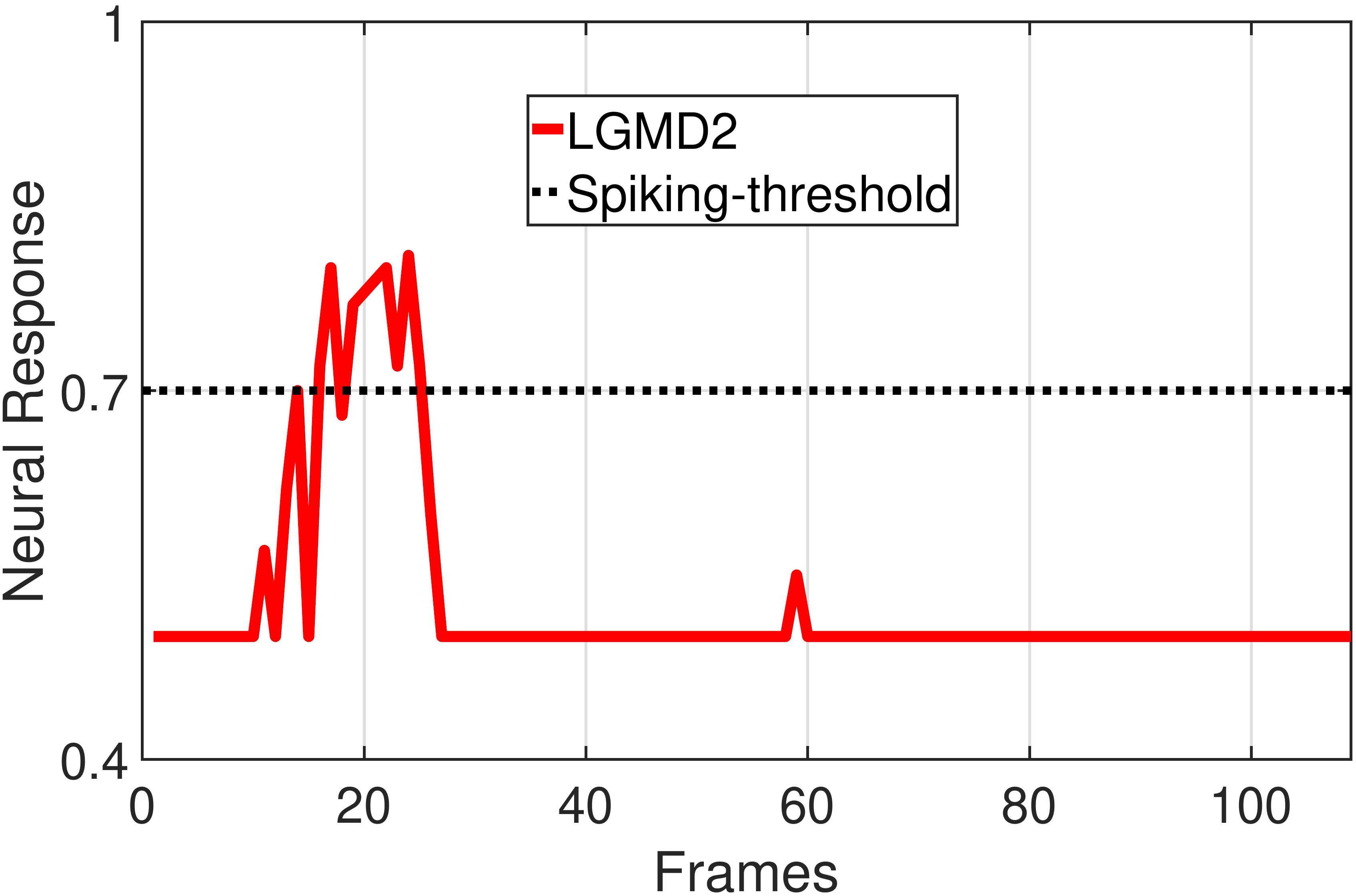}
		\label{translation-bf-lgmd2-potential}}
	\vfill
	\vspace{-0.1in}
	\subfloat{\includegraphics[width=0.24\linewidth]{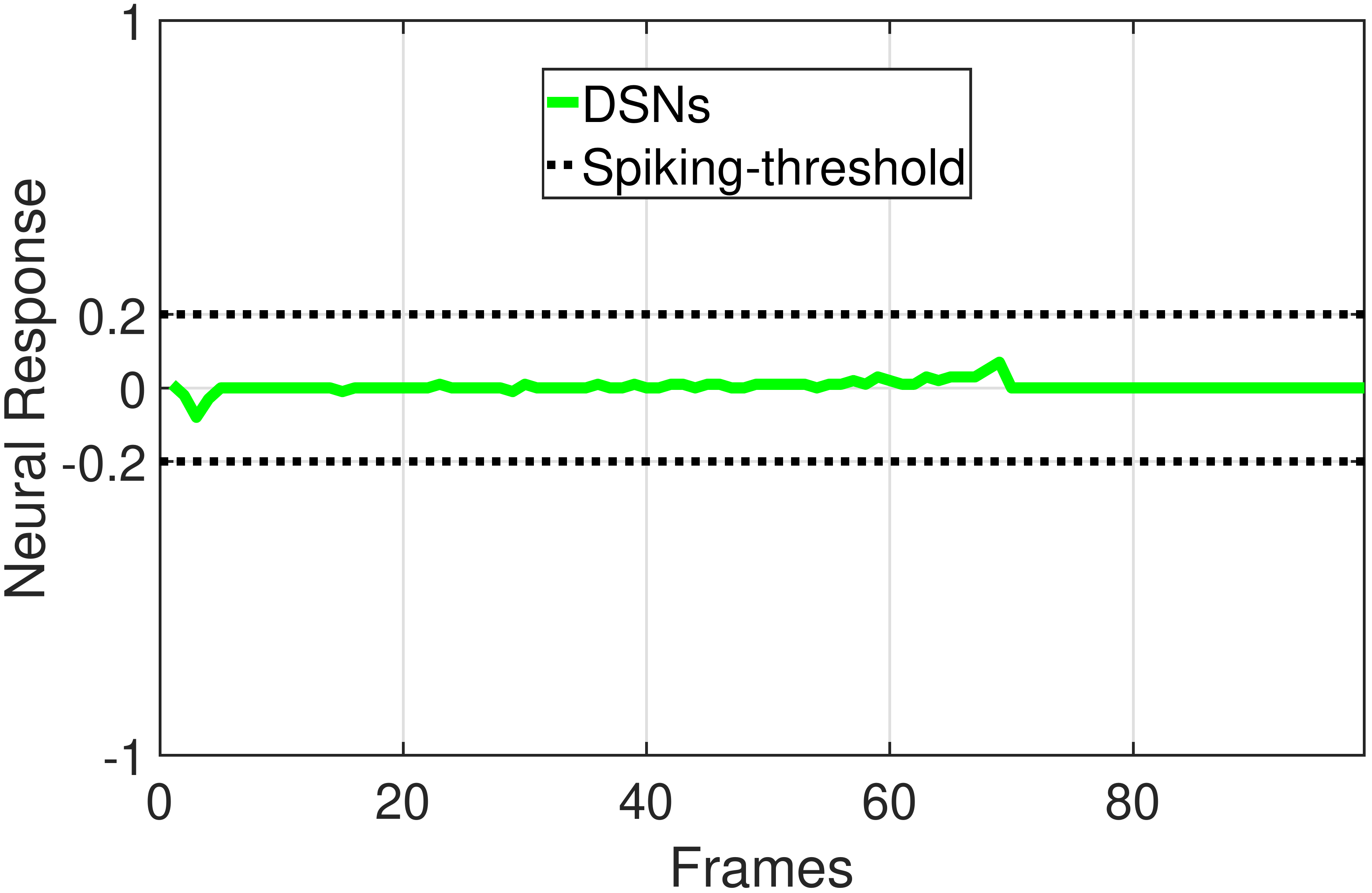}
		\label{looming-dsns-potential}}
	\hfil
	\subfloat{\includegraphics[width=0.24\linewidth]{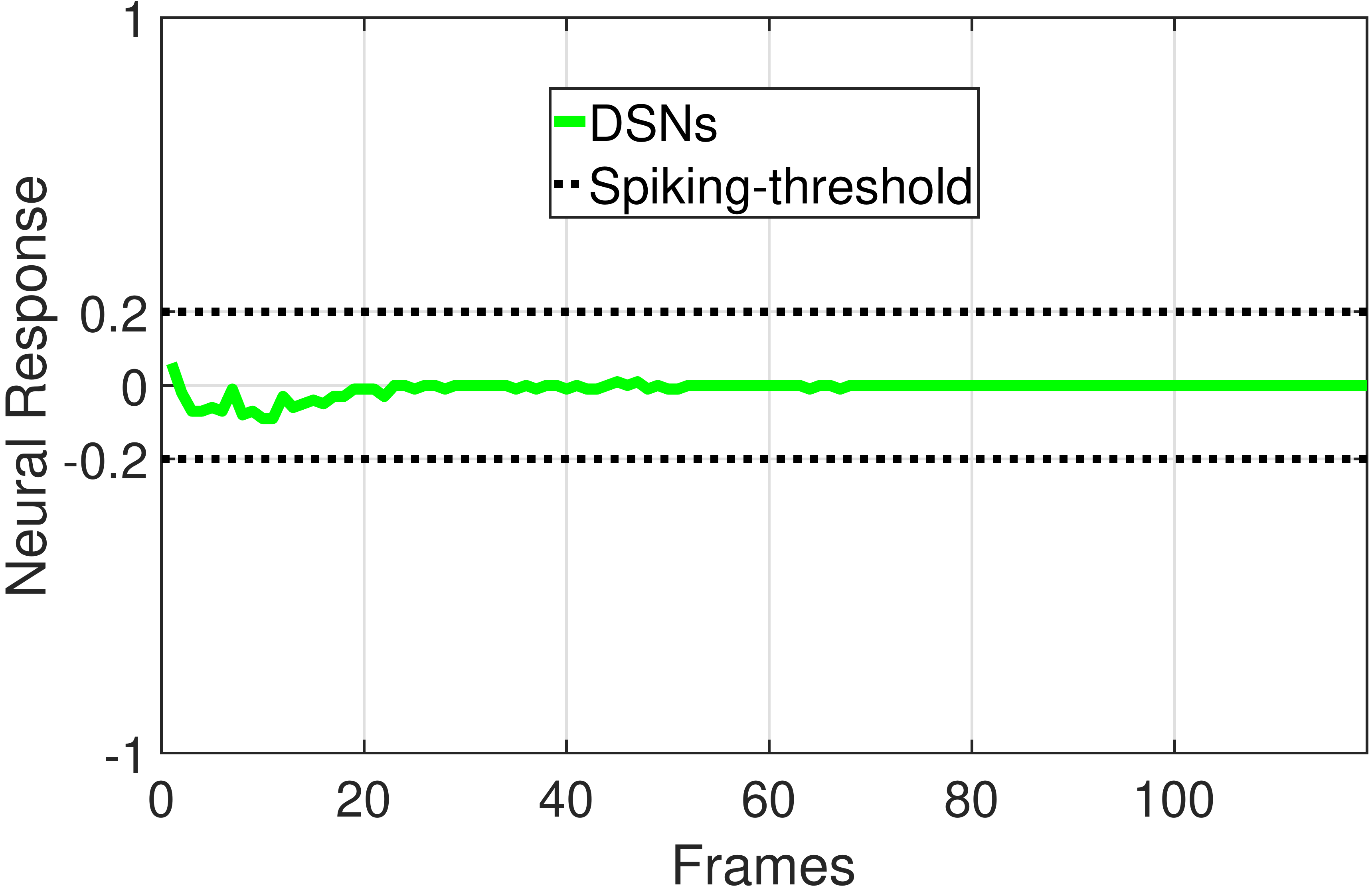}
		\label{recession-dsns-potential}}
	\hfil
	\subfloat{\includegraphics[width=0.24\linewidth]{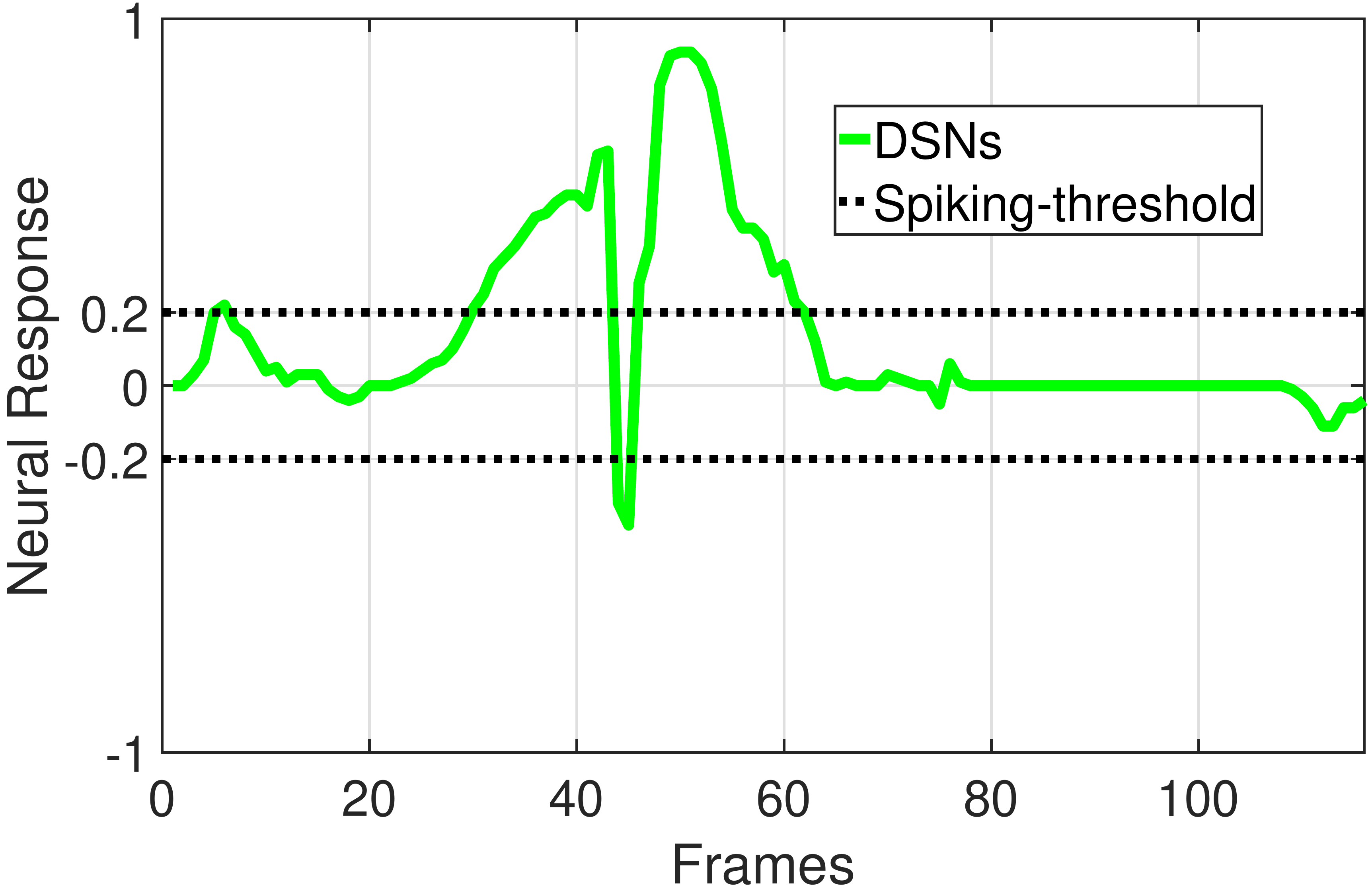}
		\label{translation-fb-dsns-potential}}
	\hfil
	\subfloat{\includegraphics[width=0.24\linewidth]{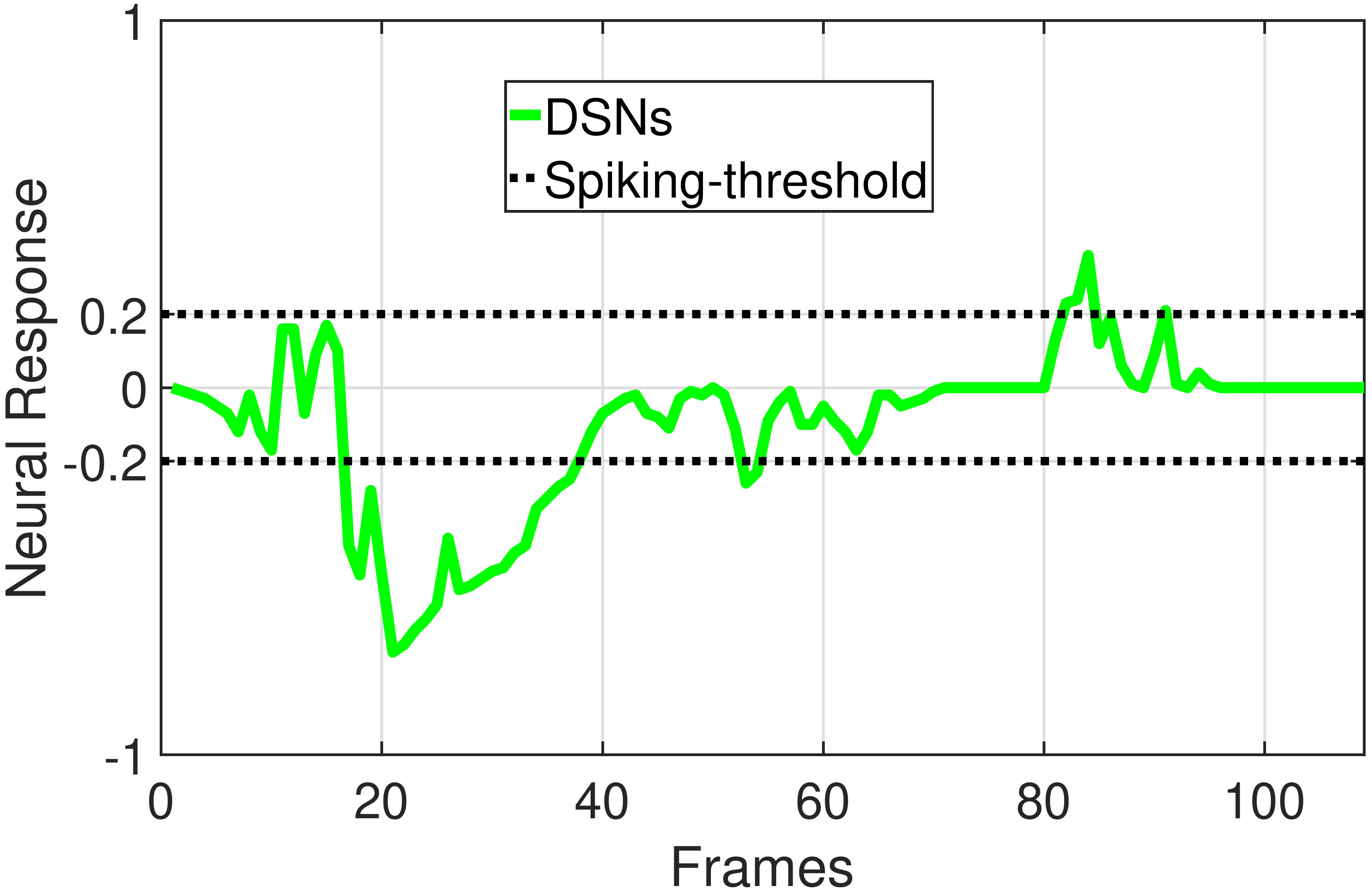}
		\label{translation-bf-dsns-potential}}
	\caption{Neural responses of four visual neurons challenged by the four basic motion patterns: looming (a), recession (b), translation-rightward (c) and translation-leftward(d). The spiking thresholds are designated by horizontally dashed lines. \textbf{The LGMDs neurons respond most strongly to the looming, whilst the LGMD2 neuron is rigorously inhibited by the recession. Conversely, the DSNs neurons are inhibited by both the looming and recession, but highly activated by the translations, that is, the DSN-R responds to the rightward translations with positive neural outputs, while the DSN-L responds to the leftward translations with negative neural outputs. The LGMDs are also activated by the translations in either directions.}}
	\label{open-robot-tests}
\end{figure*}
In the open-loop tests, we first demonstrate the neural responses of the LGMD1, the LGMD2 and the DSNs sub-systems, challenged by the four basic motion patterns, as illustrated in Fig. \ref{open-robot-tests}. We collected the model outputs including spikes and neural responses, remotely through the Bluetooth device with the motionless stimulated robot, as shown in Fig. \ref{colias-units}. Another \textit{Colias} robot was used as the visual stimulus.

The results shown in Fig. \ref{open-robot-tests} verify the complementary functionality of the LGMDs to the DSNs models. It is necessary to emphasize that the DSN-R and the DSN-L neurons are activated by positive and negative neural outputs of the DSNs sub-system, respectively. Compared to previous studies on integrating different insect visual neurons \cite{LGMD-DSNs-Collision,LGMD1-Yue2006}, for the first time the LGMD2 model is incorporated in such a synthetic neural system. The LGMD2 neuron has no response to the recession of darker objects compared to the background, which can be an ideal model for ground robotic vision system \cite{LGMD2-BMVC}. Interestingly, combining the functionality of the LGMD2 with the LGMD1 neural systems and a logical `AND' operation can well recognize the recession pattern.

\begin{figure*}[!t]
	\centering
	\subfloat{\includegraphics[width=0.23\linewidth]{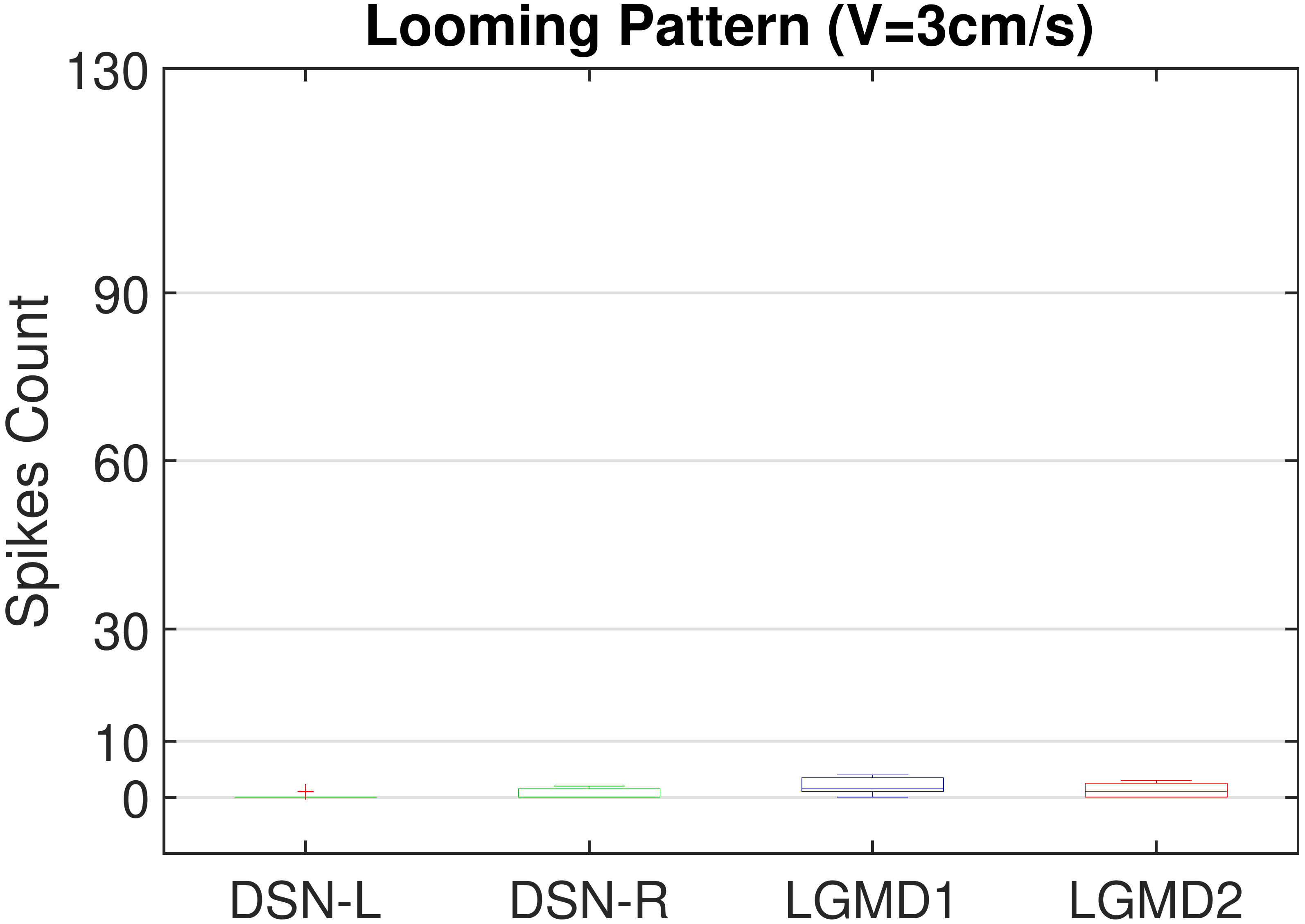}
		\label{looming-s40-box}}
	\hfil
	\subfloat{\includegraphics[width=0.23\linewidth]{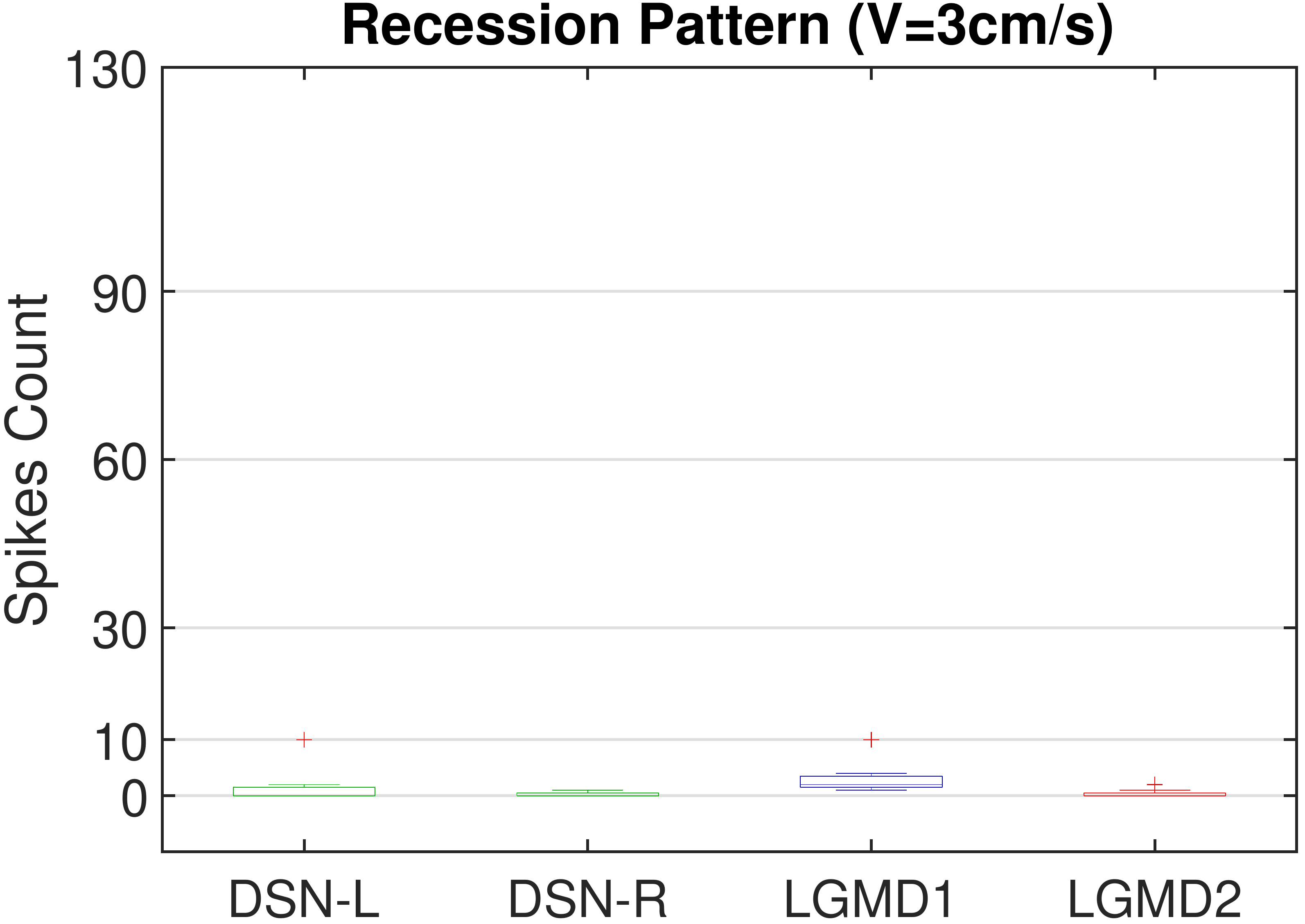}
		\label{recession-s40-box}}
	\hfil
	\subfloat{\includegraphics[width=0.23\linewidth]{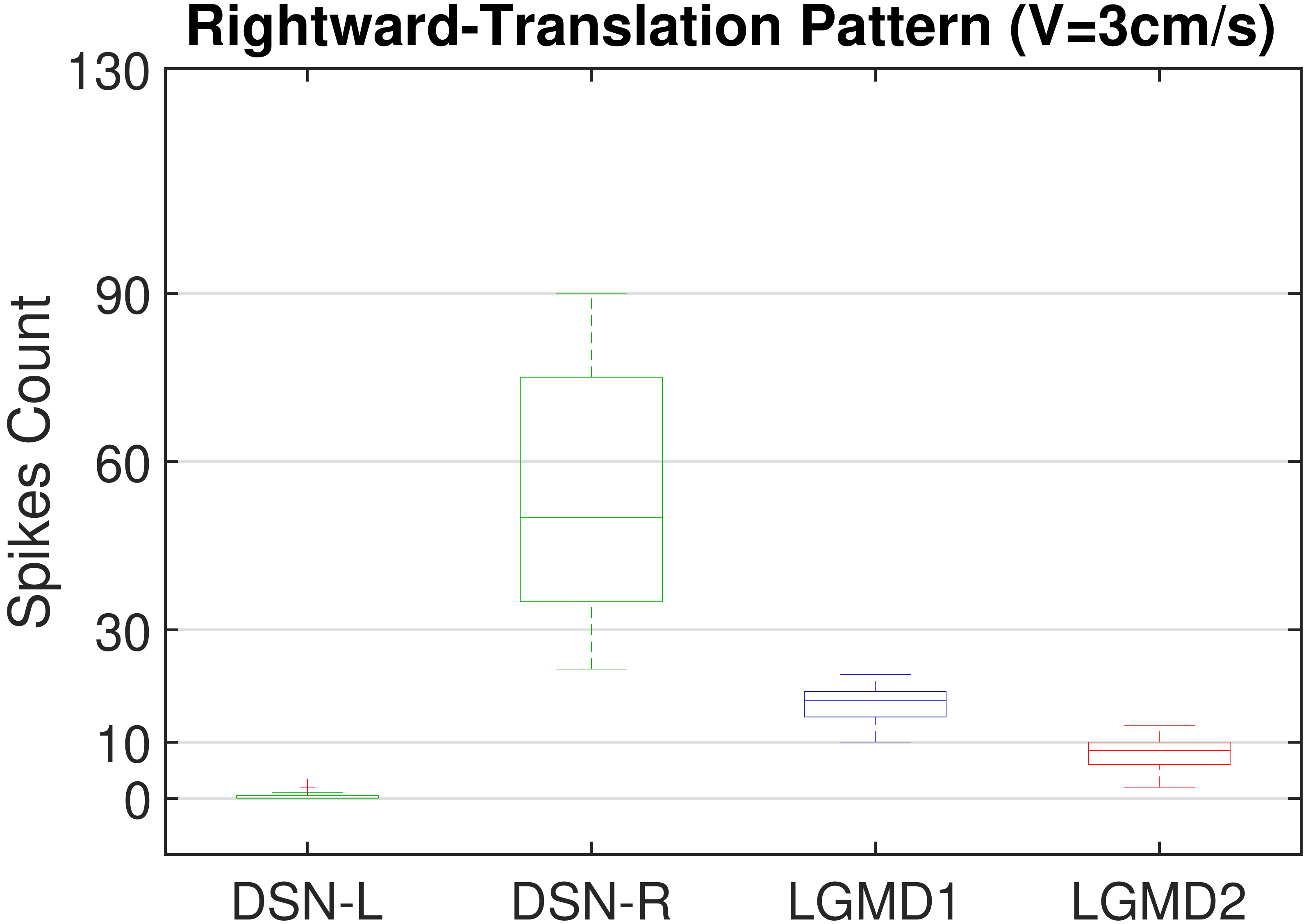}
		\label{translation-fb-s40-box}}
	\hfil
	\subfloat{\includegraphics[width=0.23\linewidth]{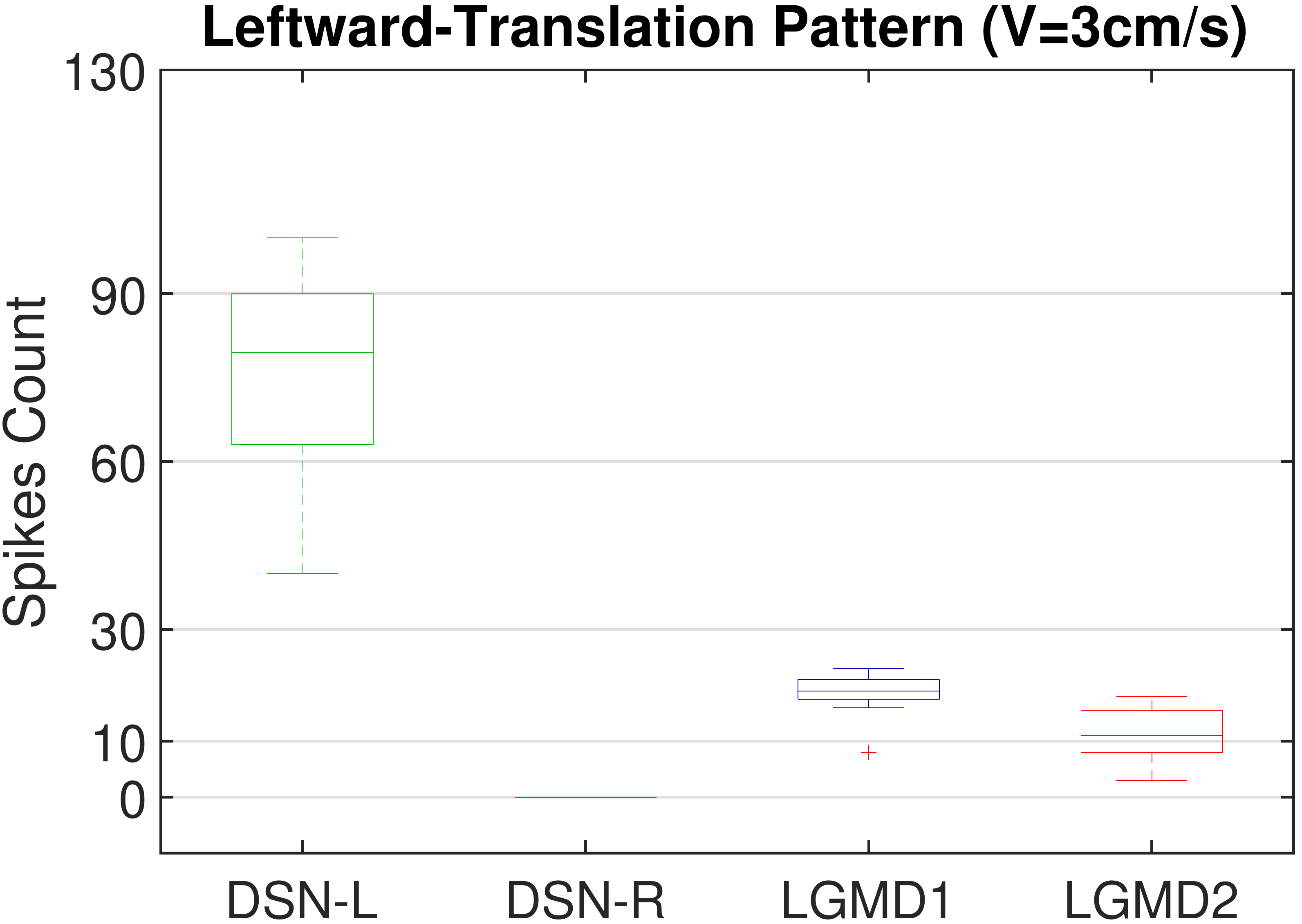}
		\label{translation-bf-s40-box}}
	\vfill
	\vspace{-0.1in}
	\subfloat{\includegraphics[width=0.23\linewidth]{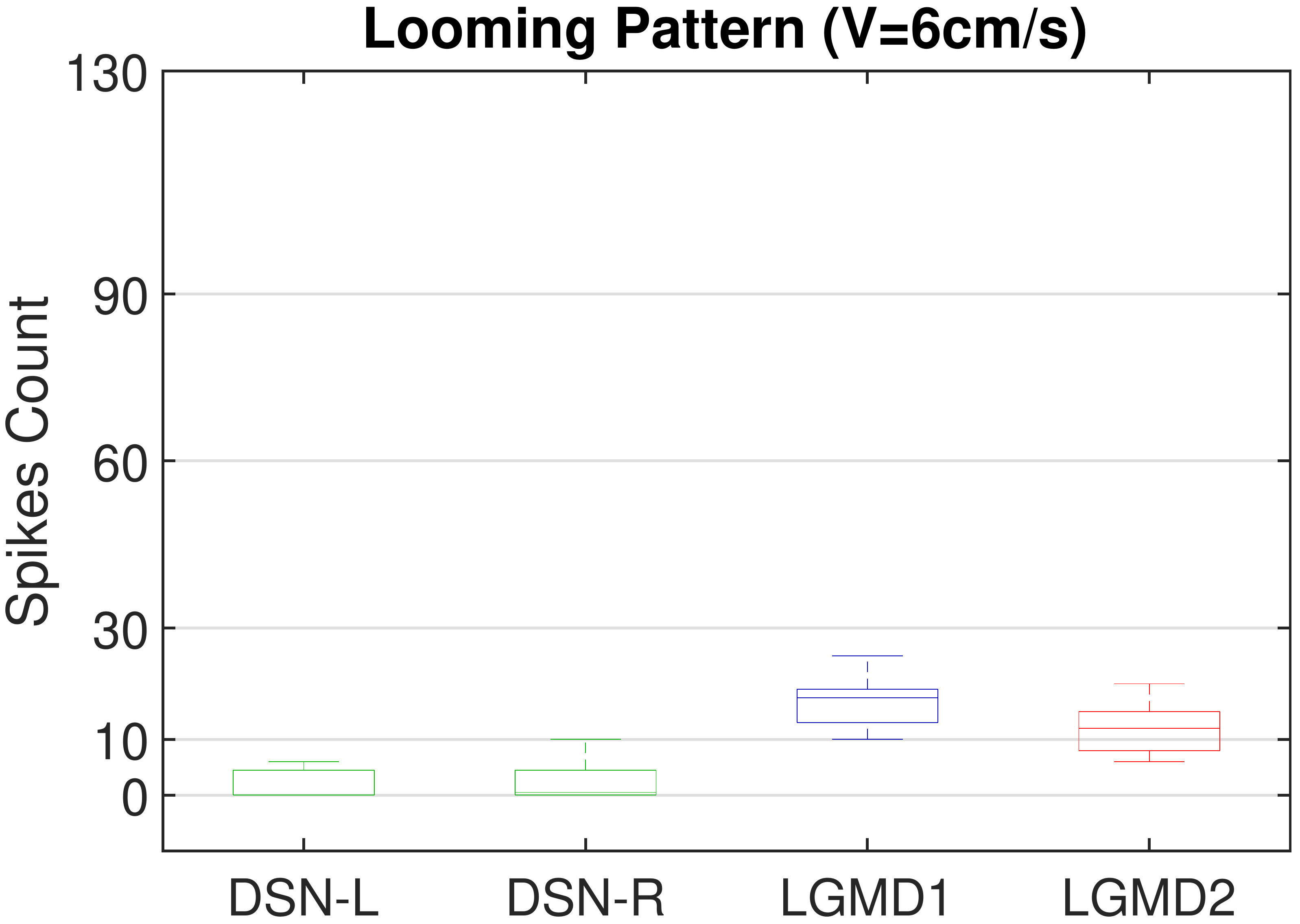}
		\label{looming-s60-box}}
	\hfil
	\subfloat{\includegraphics[width=0.23\linewidth]{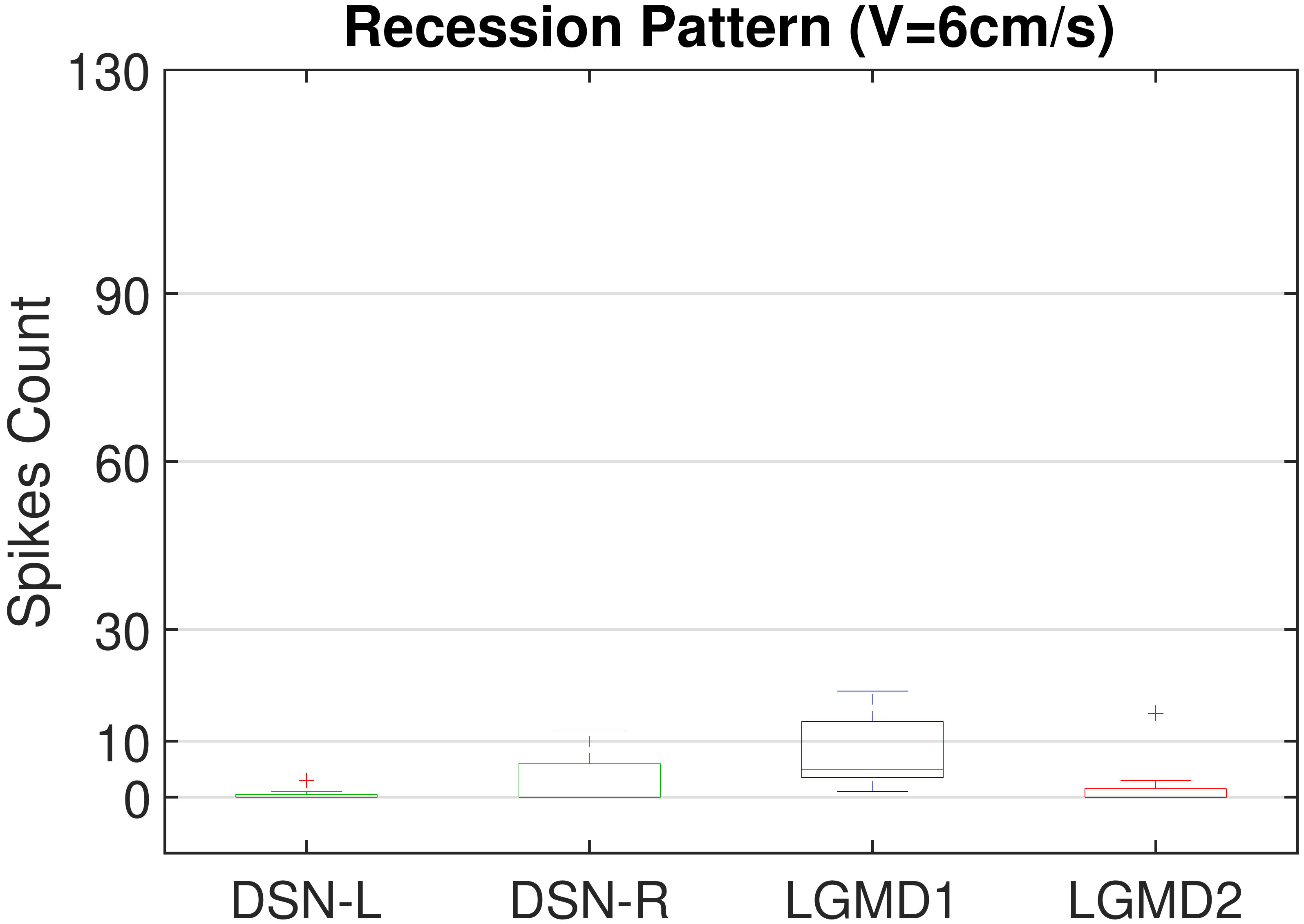}
		\label{recession-s60-box}}
	\hfil
	\subfloat{\includegraphics[width=0.23\linewidth]{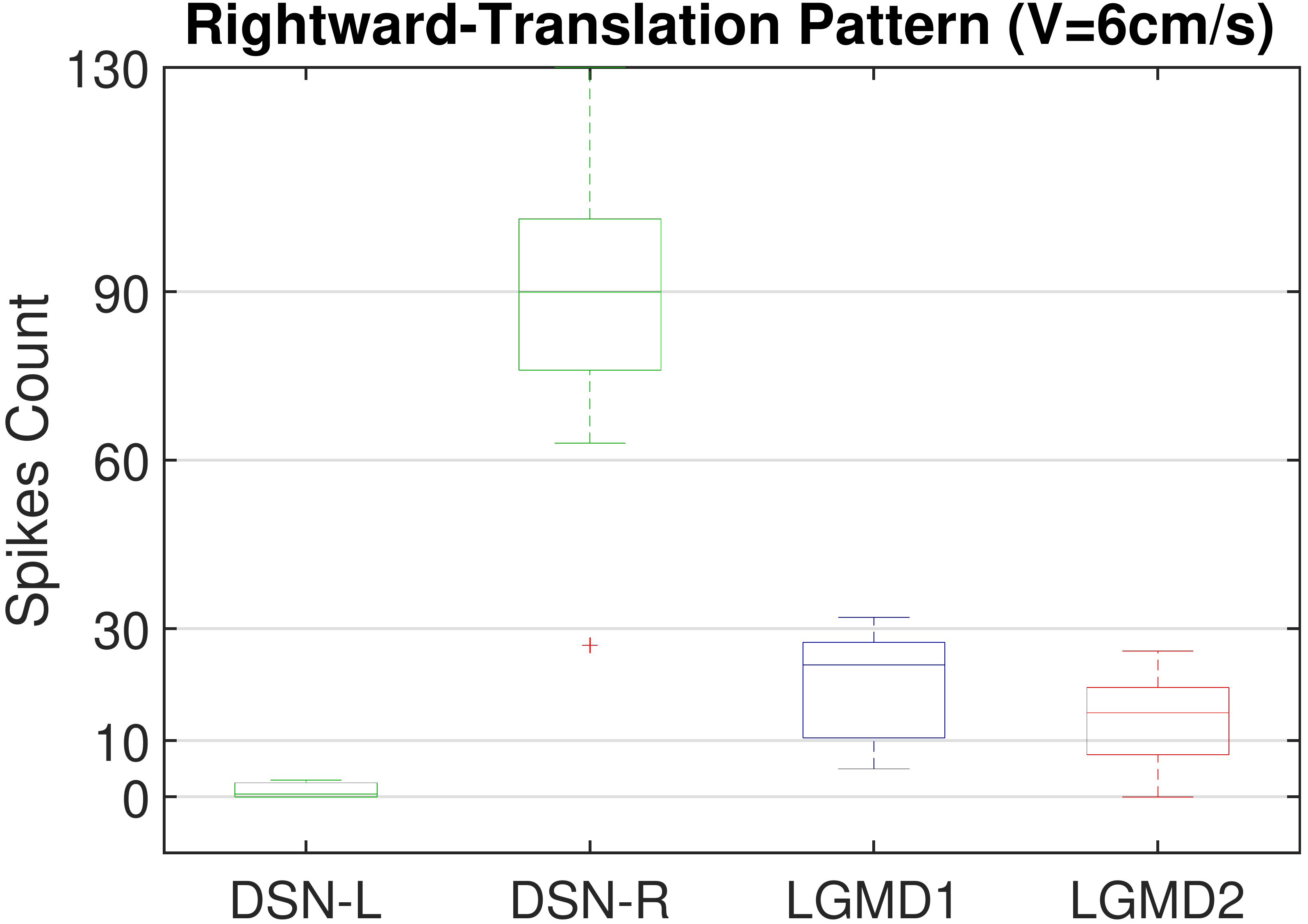}
		\label{translation-fb-s60-box}}
	\hfil
	\subfloat{\includegraphics[width=0.23\linewidth]{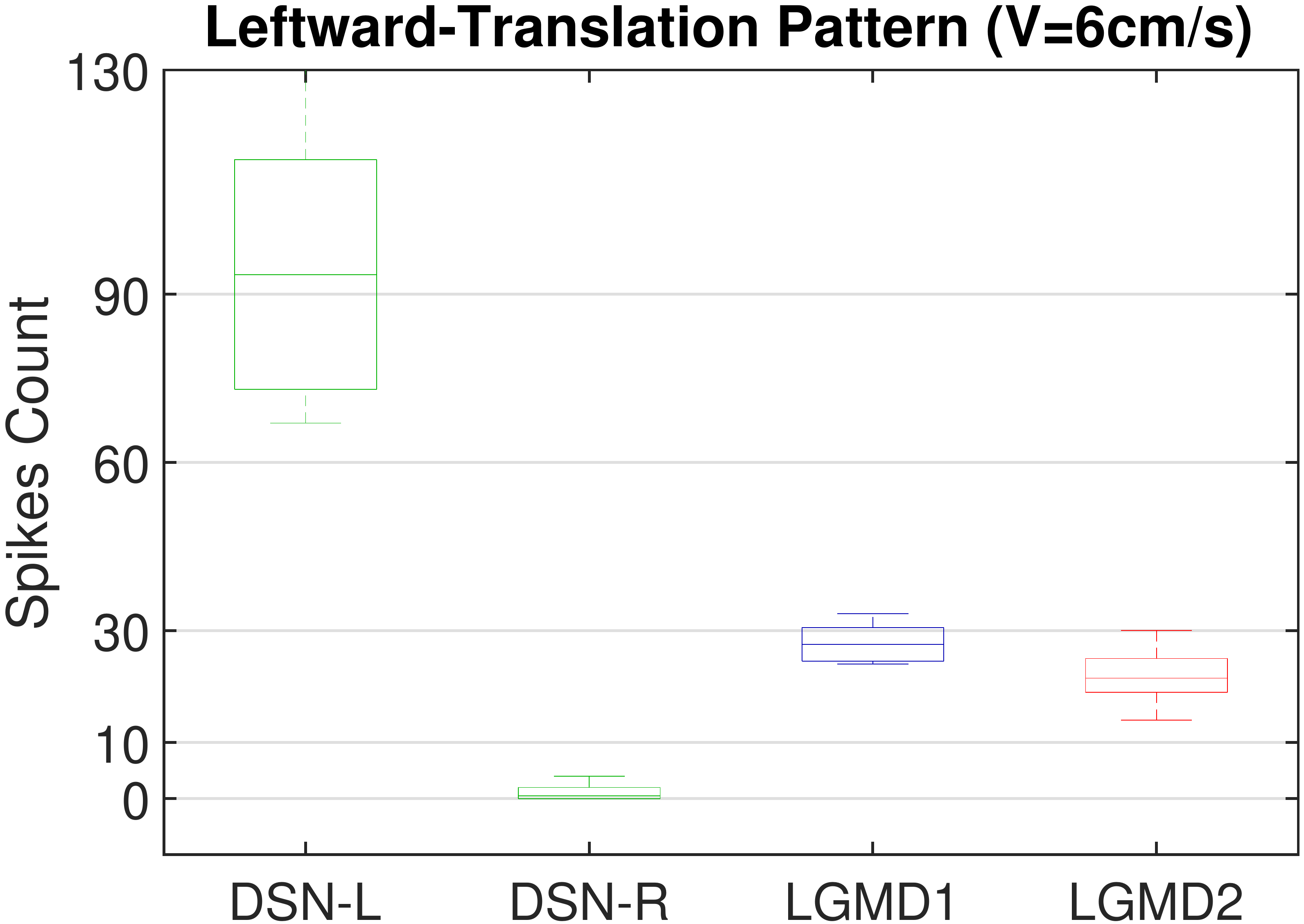}
		\label{translation-bf-s60-box}}
	\vfill
	\vspace{-0.1in}
	\subfloat{\includegraphics[width=0.23\linewidth]{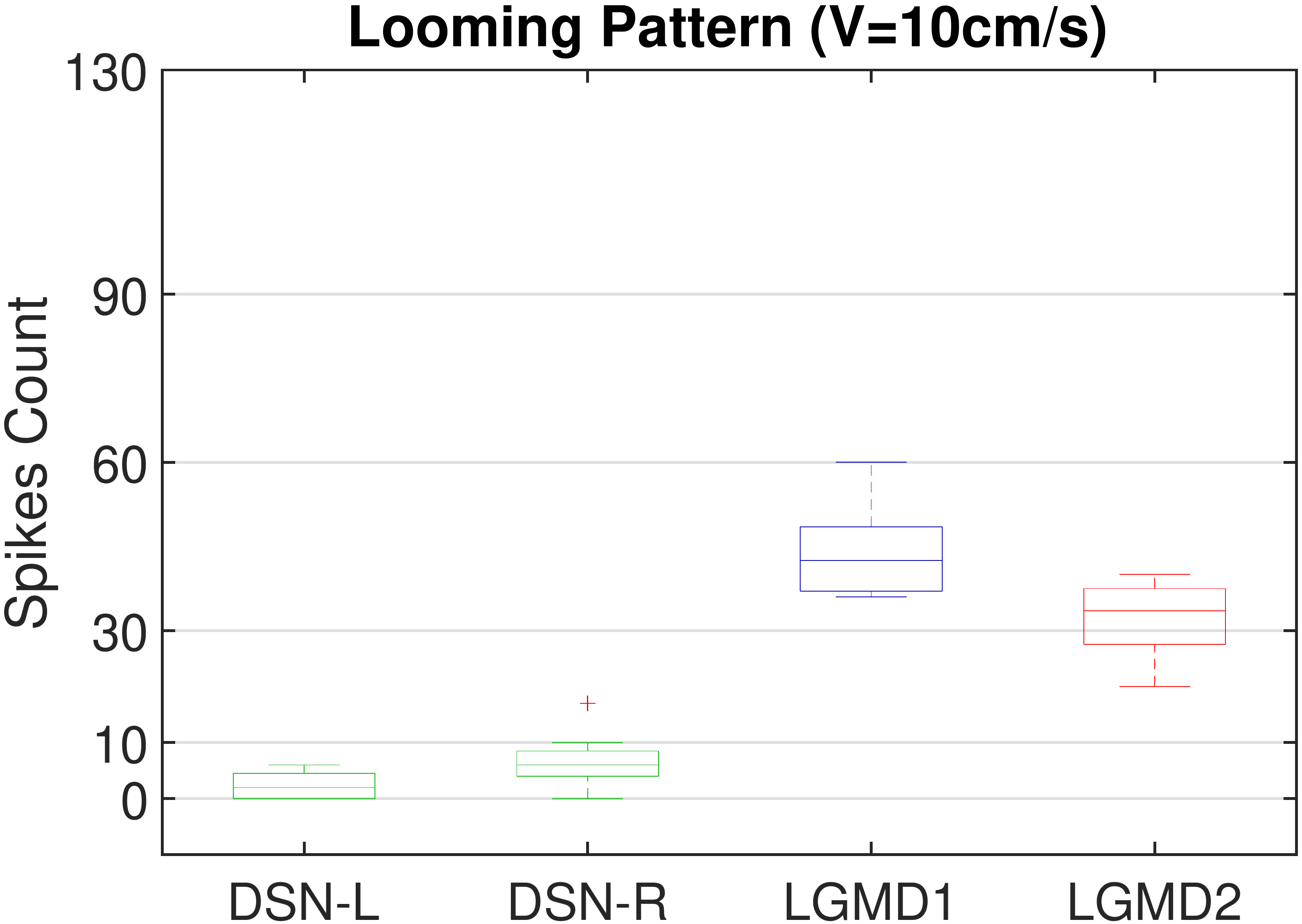}
		\label{looming-s80-box}}
	\hfil
	\subfloat{\includegraphics[width=0.23\linewidth]{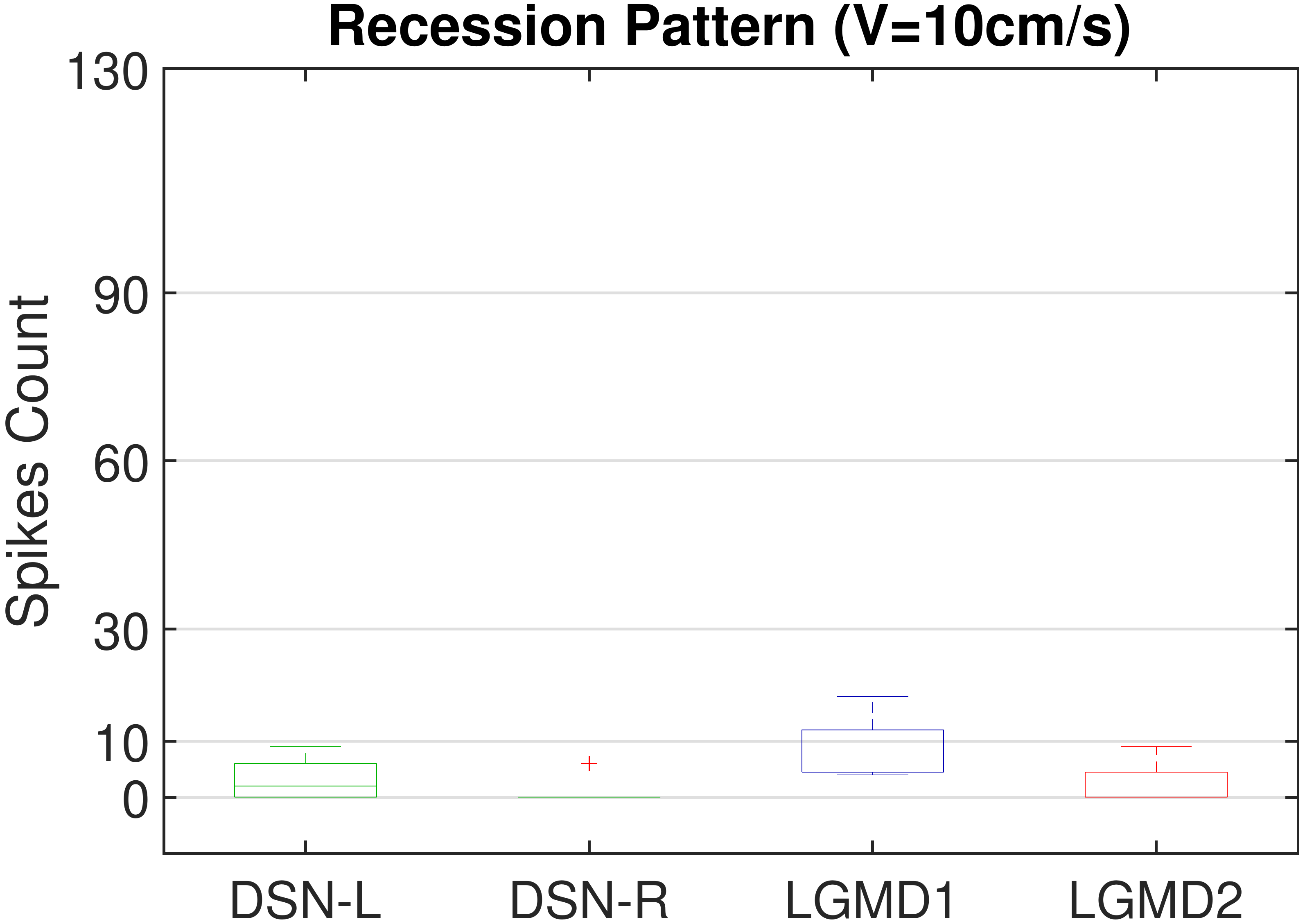}
		\label{recession-s80-box}}
	\hfil
	\subfloat{\includegraphics[width=0.23\linewidth]{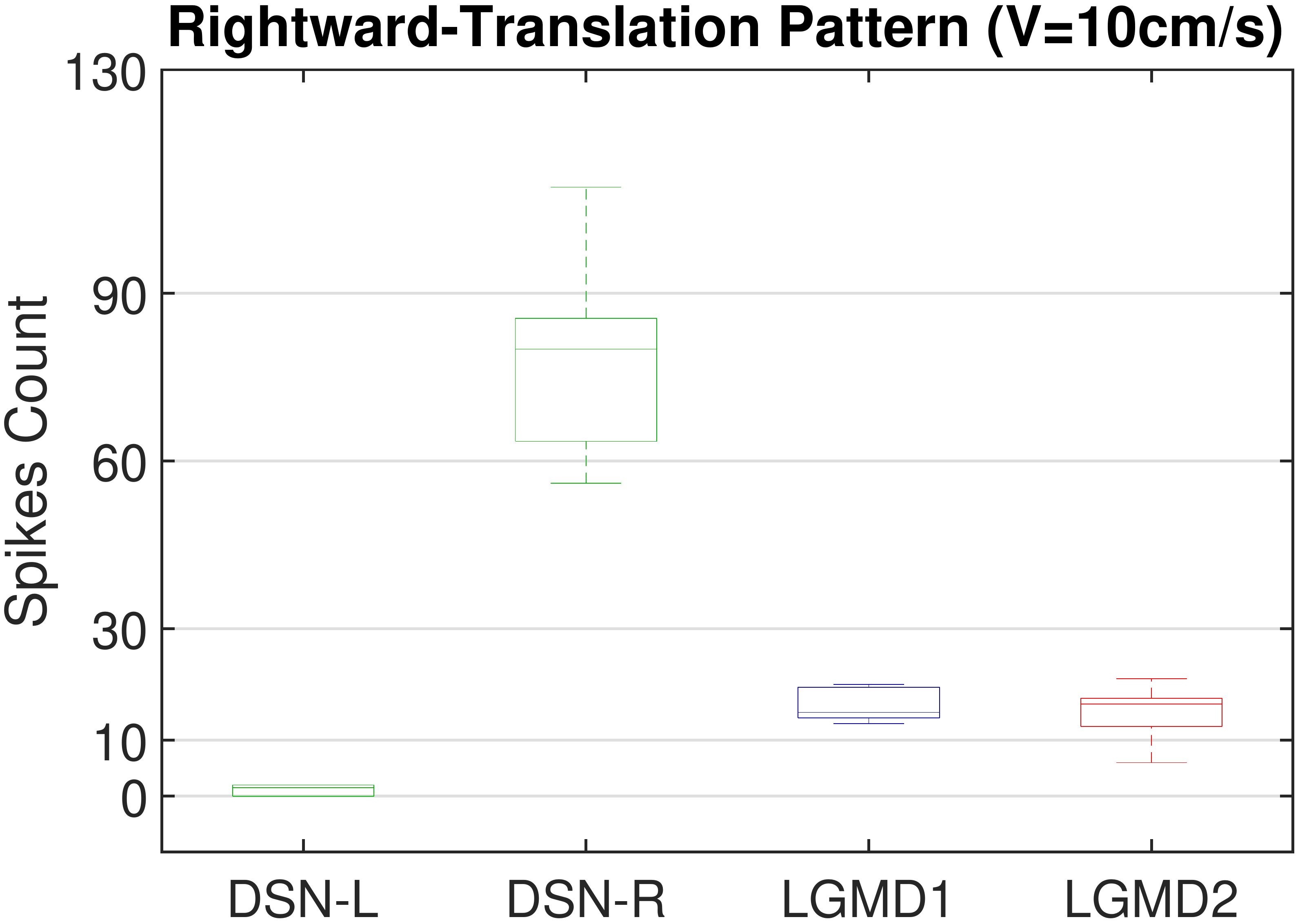}
		\label{translation-fb-s80-box}}
	\hfil
	\subfloat{\includegraphics[width=0.23\linewidth]{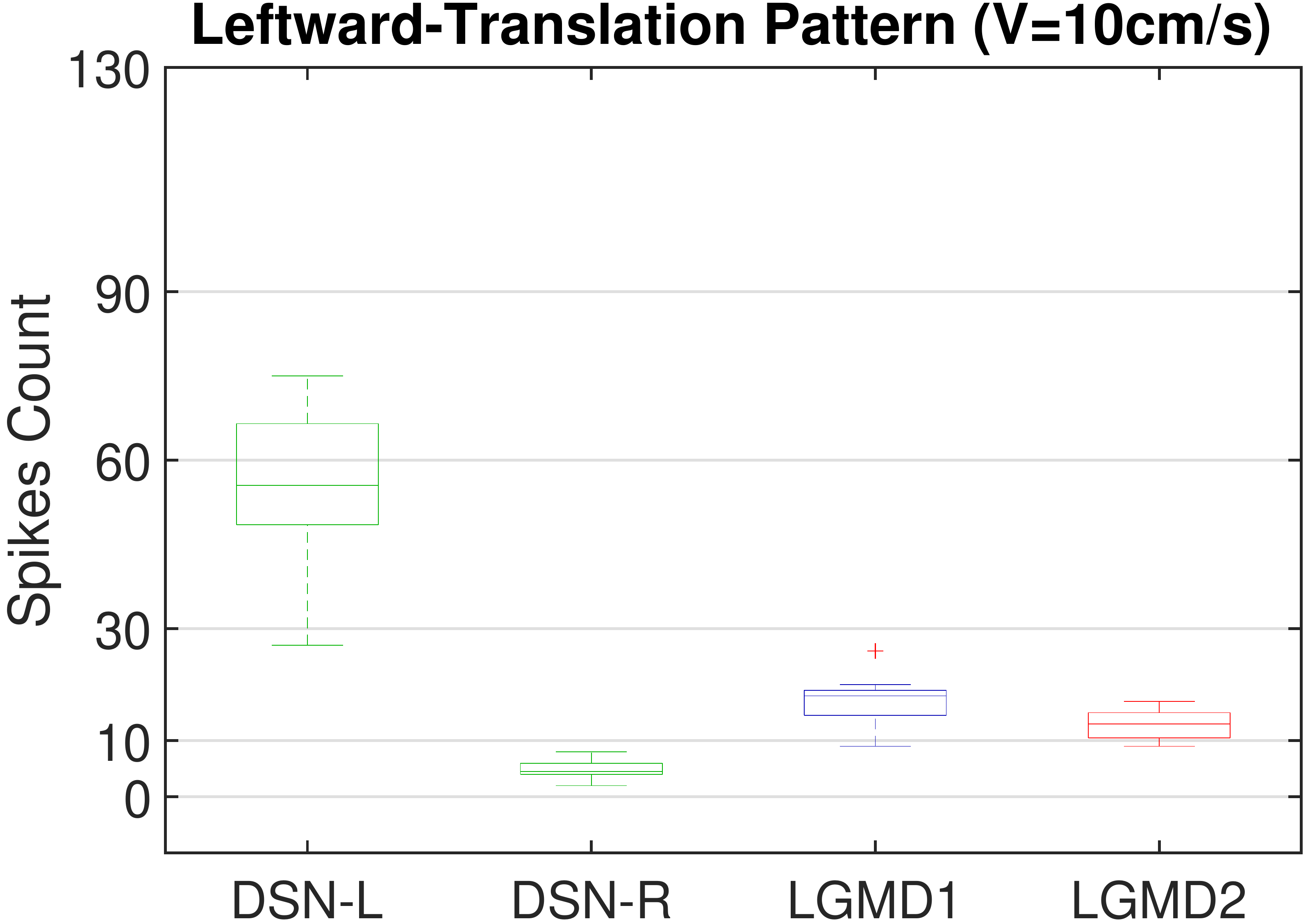}
		\label{translation-bf-s80-box}}
	\vfill
	\vspace{-0.1in}
	\subfloat{\includegraphics[width=0.23\linewidth]{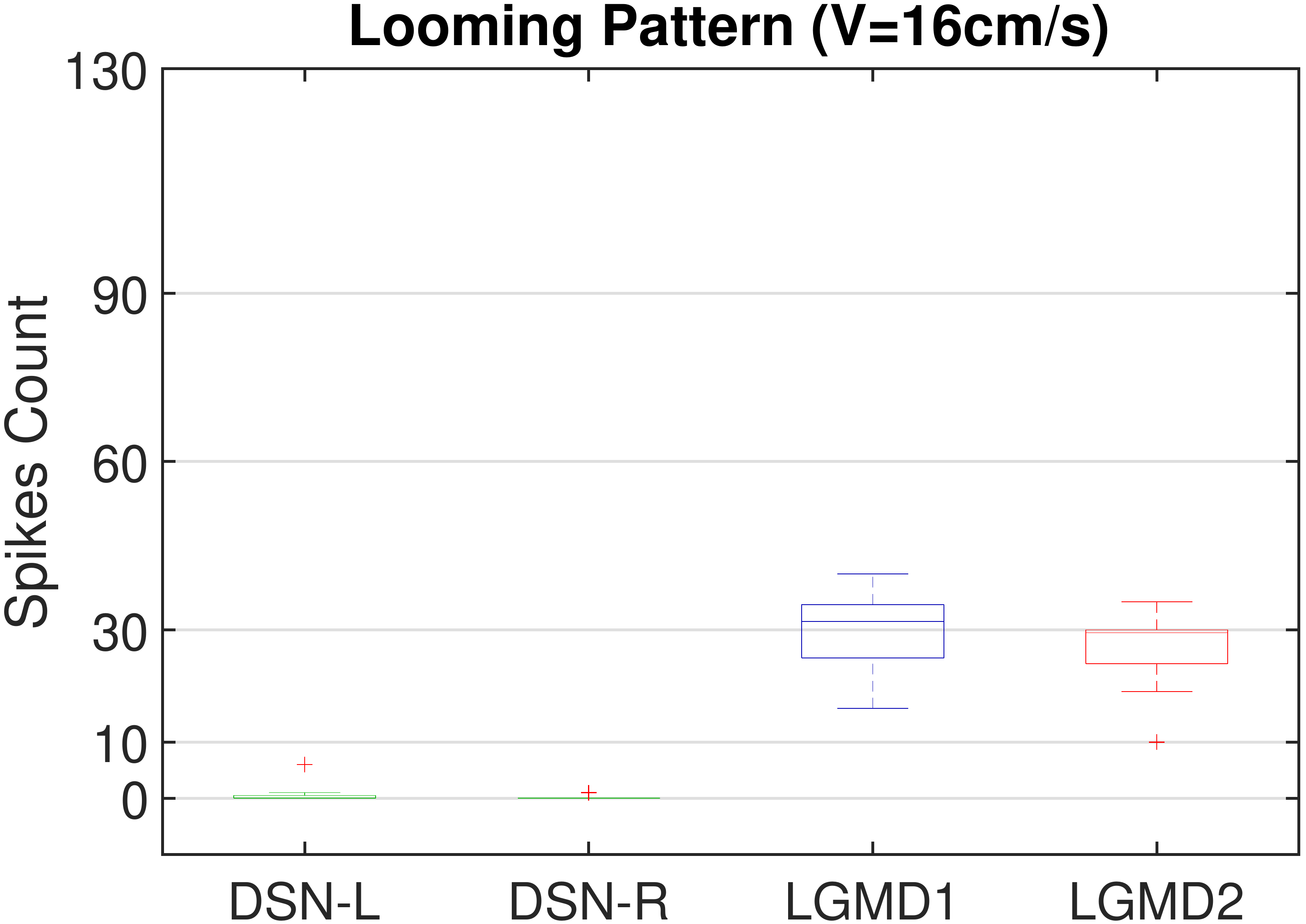}
		\label{looming-s120-box}}
	\hfil
	\subfloat{\includegraphics[width=0.23\linewidth]{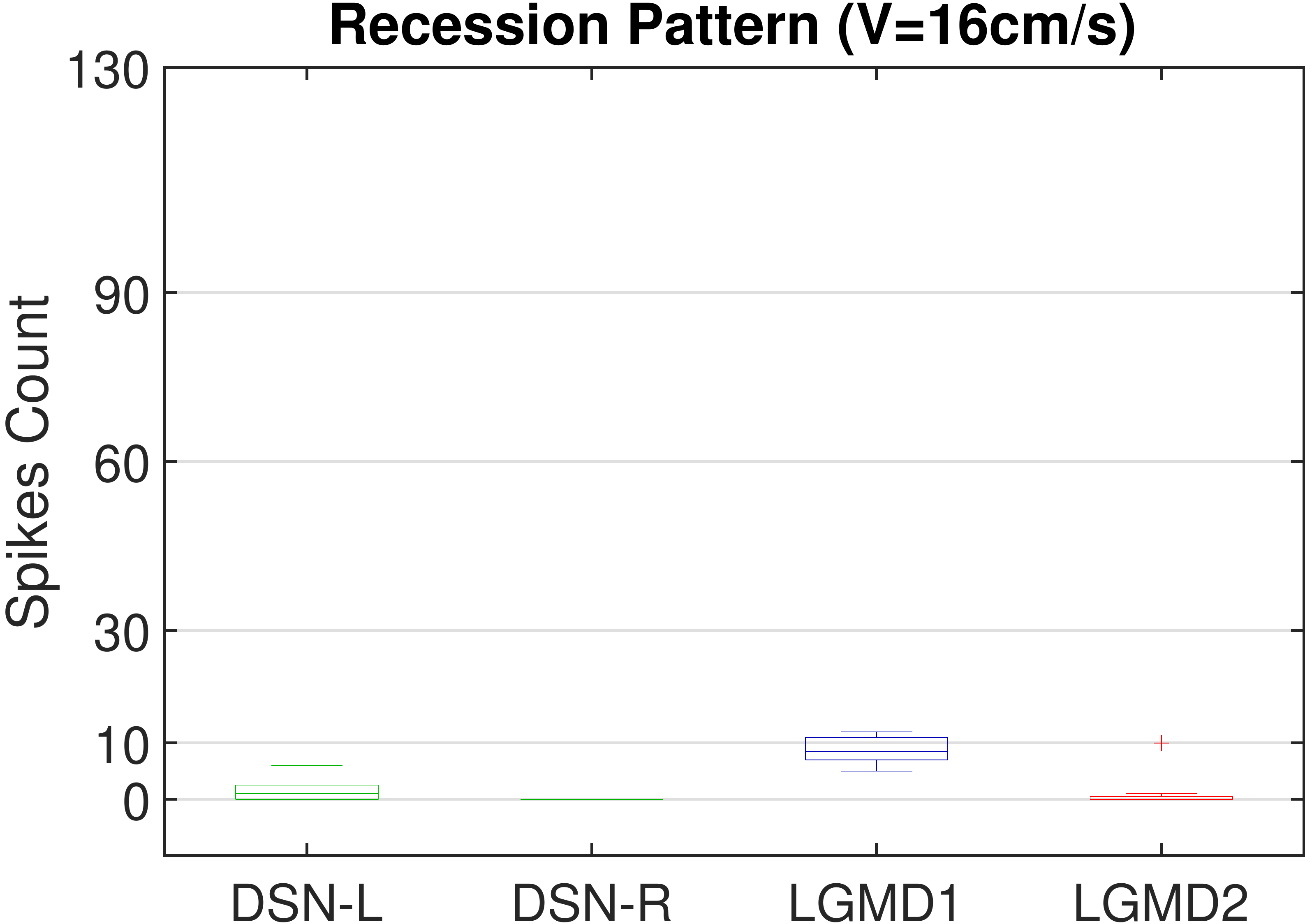}
		\label{recession-s120-box}}
	\hfil
	\subfloat{\includegraphics[width=0.23\linewidth]{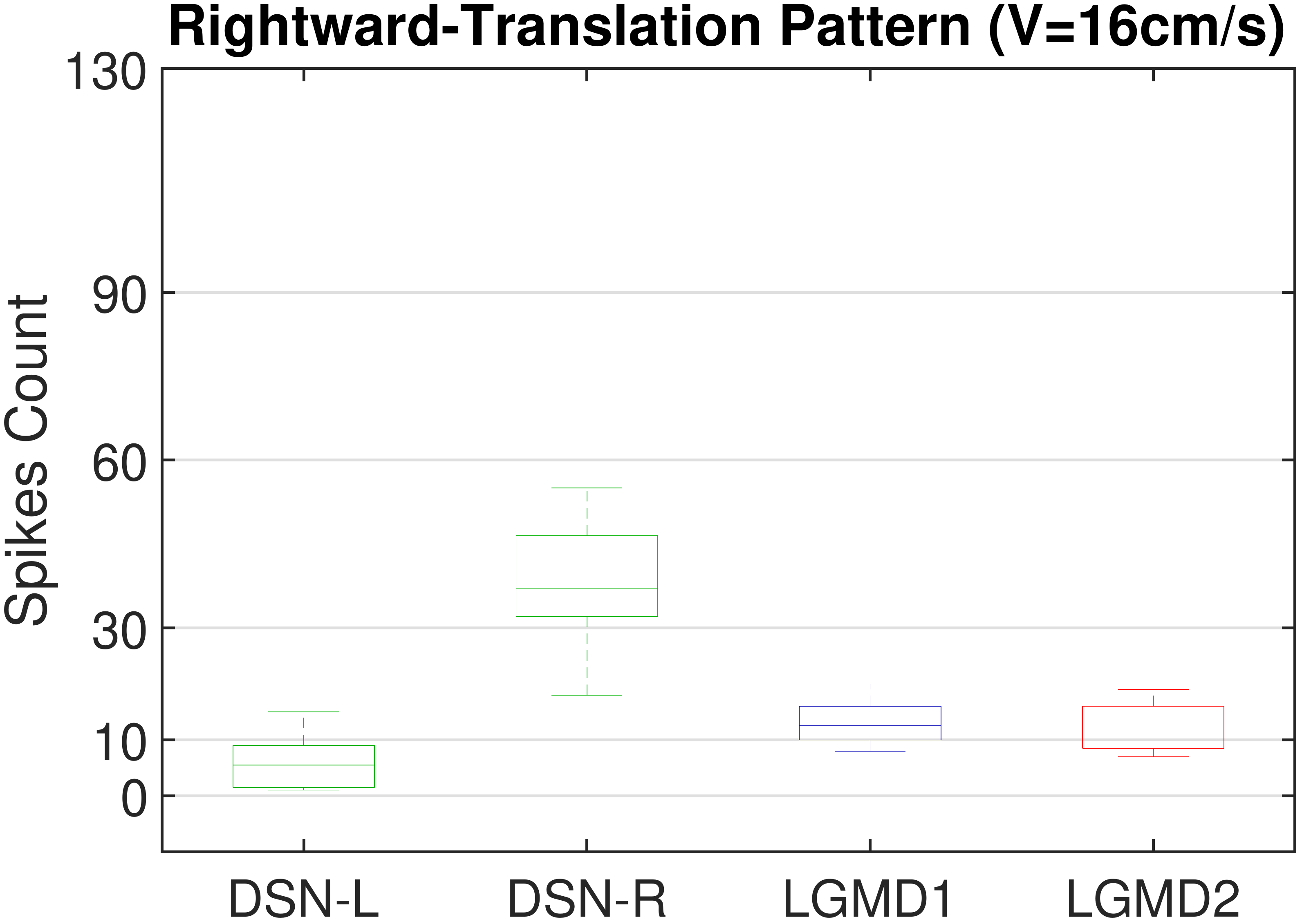}
		\label{translation-fb-s120-box}}
	\hfil
	\subfloat{\includegraphics[width=0.23\linewidth]{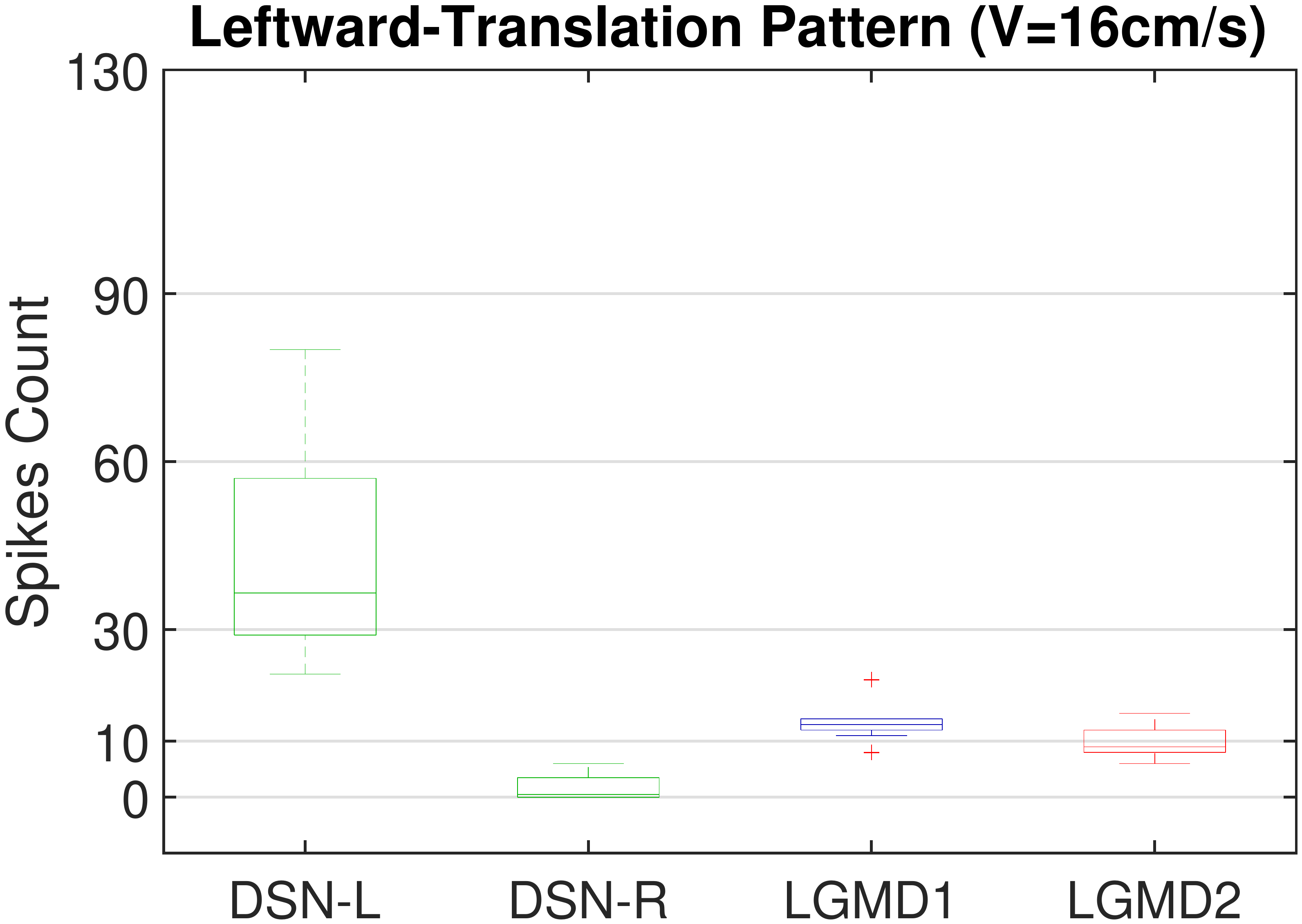}
		\label{translation-bf-s120-box}}
	\caption{Statistics of the spike frequency in the open-loop tests: the embedded vision system is tested by these four basic motion patterns, shown in Fig. \ref{open-robot-tests}, by a moving robot as the visual stimulus and at four constant speeds, respectively. Each set of movements is repeated ten times. The spikes during each course are accumulated. \textbf{The DSNs neurons represent much higher spike frequency compared to the LGMDs, challenged by the translations at all tested speeds, while they are rigorously inhibited by the looming and the recession. The LGMDs spike at high frequency by both the fast looming and the nearby translations, while they respond most strongly to the fast looming. Only the LGMD1 neuron spikes frequently by the recession.}}
	\label{open-robot-statistics}
\end{figure*}

Moreover, we demonstrate the effects of each spiking neuron on recognizing different motion patterns. Intuitively, the statistical results in Fig. \ref{open-robot-statistics} demonstrate that the DSNs neurons spike at much higher rate than the LGMDs neurons, when challenged by translations from slow to fast speeds, respectively. More specifically, the DSN-R and the DSN-L rigorously spike at high rate by the rightward and leftward translations, respectively, even tested by very fast movements. On the other hand, the LGMDs are activated by fast approaching and also the nearby translations, while both are not activated by the looming at very slow speed ($3cm/s$ in our case). The LGMD1 normally spikes at higher rate than the LGMD2. However, the LGMD2 remains quiet during the robot recession, but the LGMD1 is not. In our switch function of the embedded vision system, the activation of DSNs will rigorously inhibit the LGMDs, and vice versa. Therefore, the results reveal great potential in enhancing the collision selectivity and adding in new object tracking behavior in dynamic robot scenes.

\begin{figure}[t]
	\centering
	\subfloat[experimental setting]{\includegraphics[width=0.23\textwidth]{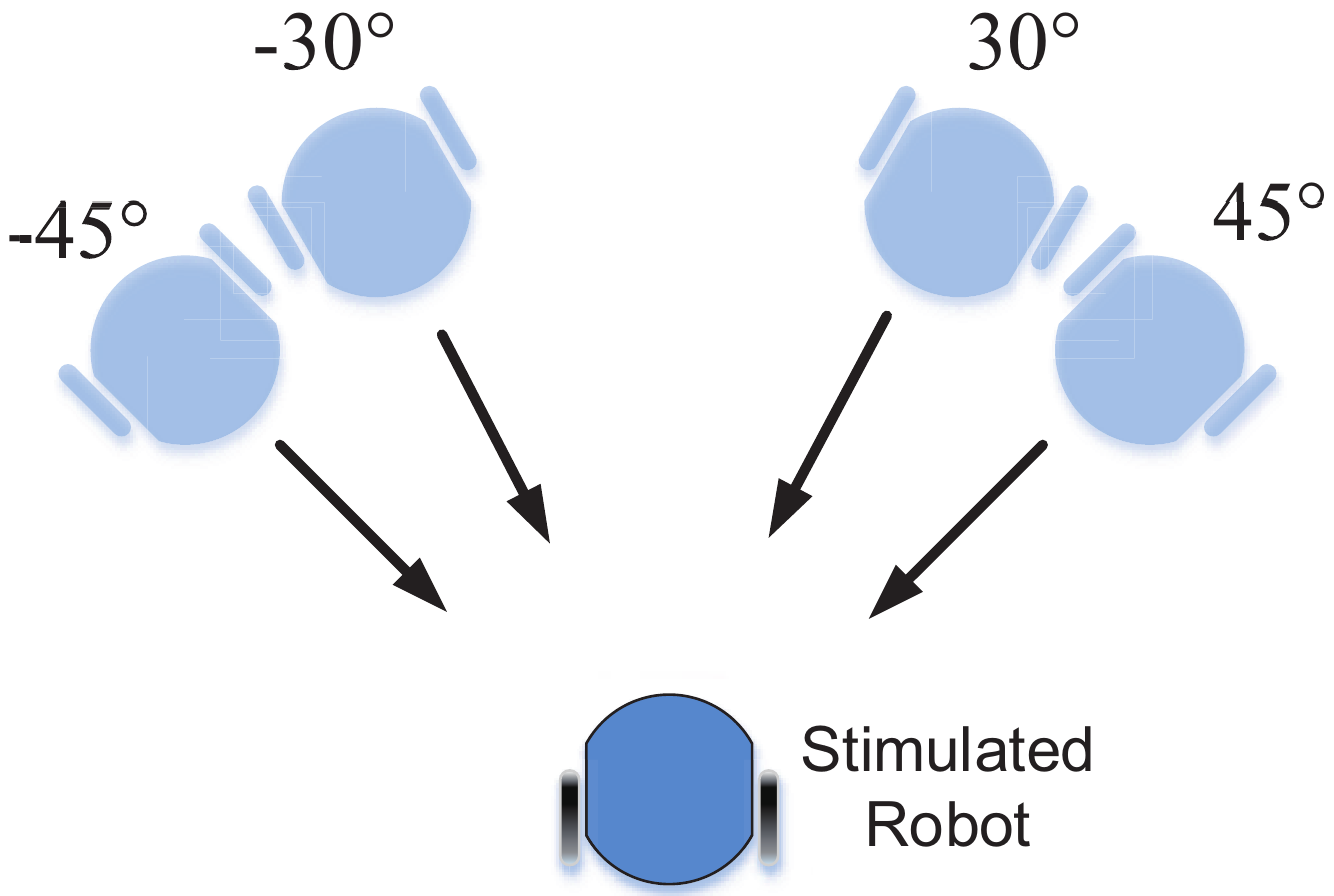}
		\label{aa-setup}}
	\hfil
	\subfloat[spike rate]{\includegraphics[width=0.21\textwidth]{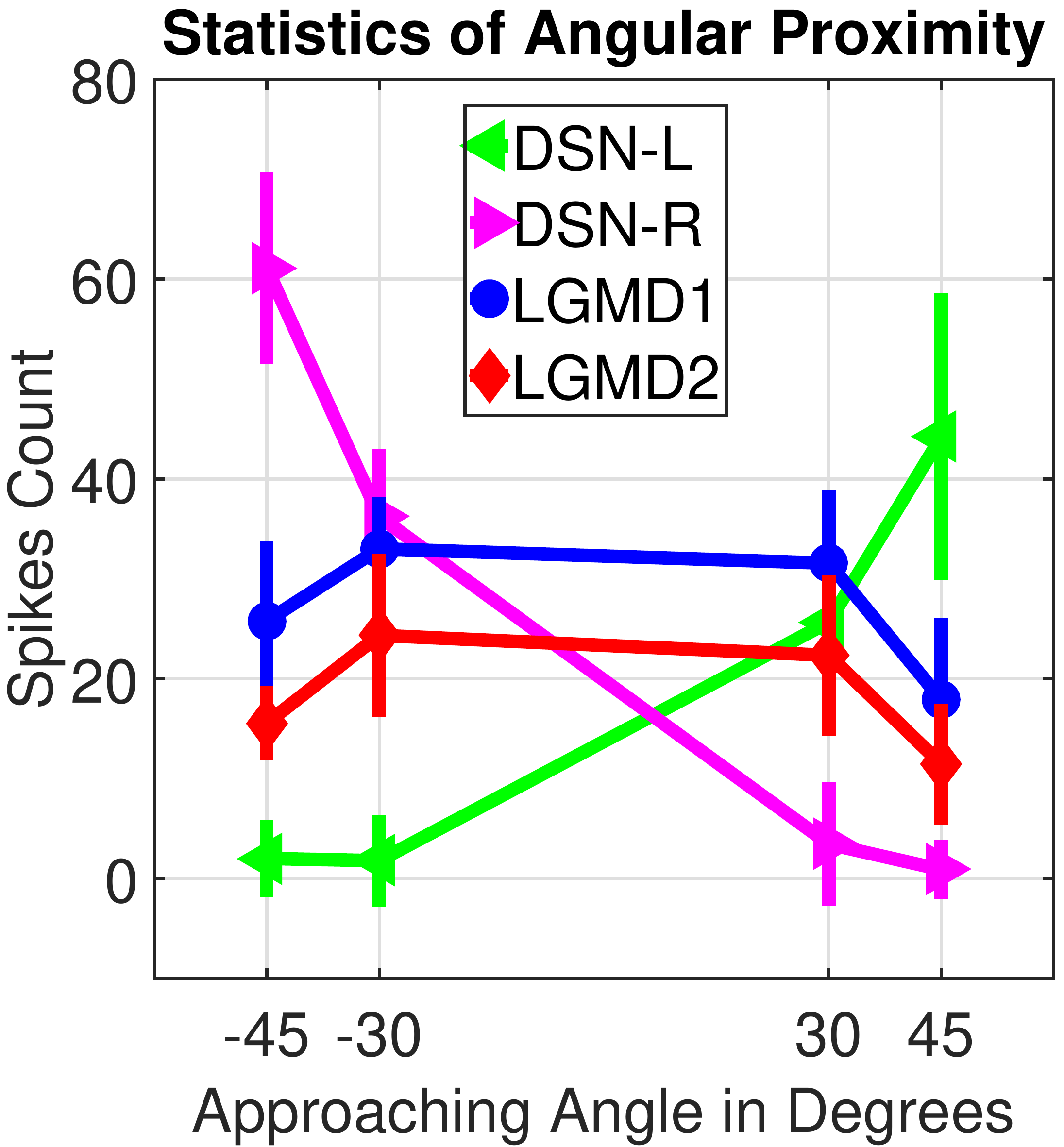}
		\label{aa-statistics}}
	\caption{Statistics of the robot angular approaching tests.}
	\label{angular-approach-tests}
\end{figure}

Furthermore, from our previous studies \cite{IROS-LGMDs}, we observed that the movements of angular approaching frequently happens in dynamic robot scenes. These visual stimuli usually activate the LGMDs neurons and trigger the collision avoidance behaviors \cite{IROS-LGMDs}. In this biorobotic study, we also investigate the influence of angular looming on the motion pattern recognition. The experimental setting is shown in Fig. \ref{aa-setup}. Each angular looming was repeated ten times. Similarly to the statistical tests in Fig. \ref{open-robot-statistics}, the spikes count corresponds to the spike frequency during each motion course with an identical speed. The statistical results in Fig. \ref{aa-statistics} demonstrate that the DSNs sub-system is more sensitive to the angular looming from large angles than the LGMDs sub-system, corresponding to the results in Fig. \ref{open-robot-statistics}. Concretely, the angular approaching from the left side of the view field gives rise to a rightward translation pattern, so that highly activated the DSN-R. The angular approaching from the right side of view thus corresponds to a leftward translation pattern, which is attractive to the DSN-L. As a result, the proposed synthetic neural system shapes the collision selectivity of the LGMDs to direct or small angular looming only.

\subsection{Arena Tests}
\label{arena-tests}
\begin{table}[t]
	\caption{Success Rates of Looming Recognition in Arena Tests}
	\centering
	\begin{tabular}{|l|l|l|}
		\toprule
		\multicolumn{3}{c}{\textbf{Events}: Colliding with Robots/Peripheries(CwR/CwP)}\\
		\multicolumn{3}{c}{Avoiding Looming/Translating Robots(ALR/ATR) or Peripheries(AP)}\\
		\multicolumn{3}{c}{SR1=AP/(AP+CwP)$\cdot$100\%, SR2=ALR/(ALR+ATR+CwR)$\cdot$100\%}\\
		\cmidrule{1-3}
		Neural Systems (4-Robots Scenes)&SR1	   			&SR2\\
		\cmidrule{1-3}
		LGMD2							&96.7\%	   			&80.0\%\\
		LGMD1 \& LGMD2					&88.1\%	   			&73.9\%\\
		\textbf{LGMDs \& DSNs}			&\textbf{90.3\%}	&\textbf{87.3\%}\\
		\cmidrule{1-3}
		Neural Systems (7-Robots Scenes)&SR1	   			&SR2\\
		\cmidrule{1-3}
		LGMD2							&95.0\%	   			&75.2\%\\
		LGMD1 \& LGMD2					&81.7\%	   			&67.8\%\\
		\textbf{LGMDs \& DSNs}			&\textbf{83.4\%}	&\textbf{90.6\%}\\
		\bottomrule
	\end{tabular}
	\label{sr-table}
\end{table}
In the second part of robot tests, we investigate the effectiveness of the proposed method in dynamic robot scenes. We designed arena tests and compared its performance with two former studies: a neural system with the LGMD2 neuron only \cite{LGMD2-BMVC}, and a hybrid model with both LGMDs neurons \cite{IROS-LGMDs}, for the purpose of examining the enhanced looming selectivity of this synthetic neural system, which was demonstrated in the open-loop tests. With these three models, two density of multiple \textit{Colias} robots moved concurrently in the arena at two tested speeds ($6$ and $10$cm/s), respectively and each lasting for one hour. We recorded the arena tests using a top-down camera and applied a robot localization system \cite{Colias-localization}.

\begin{figure}[t]
	\centering
	\subfloat[collision avoidance]{\frame{\includegraphics[width=0.45\textwidth]{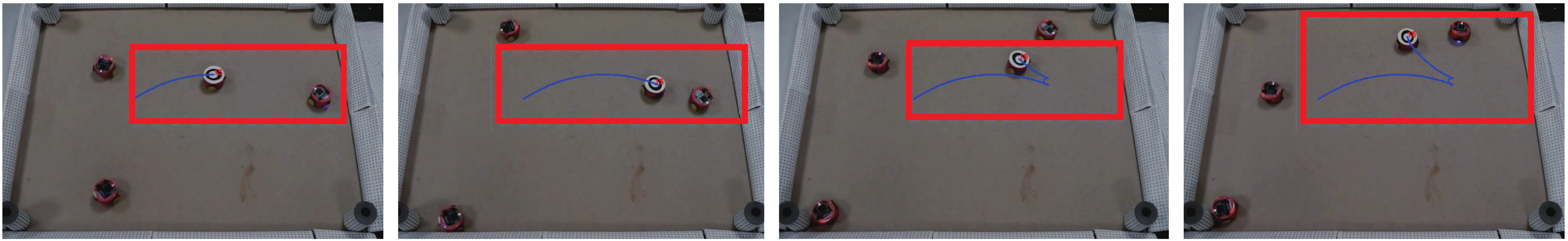}}
		\label{r2r1}}
	\vfill
	\vspace{-0.1in}
	\subfloat[tracking a translating robot]{\frame{\includegraphics[width=0.45\textwidth]{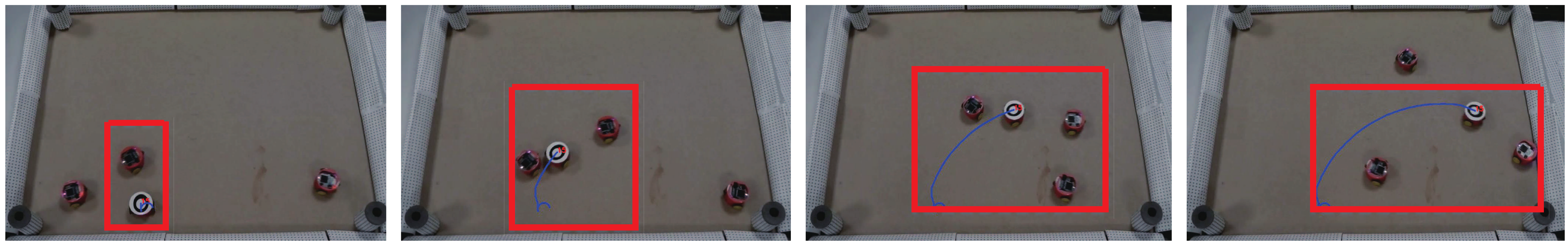}}
		\label{r2r2}}
	\vfill
	\vspace{-0.1in}
	\subfloat[non-collision with a translating robot]{\frame{\includegraphics[width=0.45\textwidth]{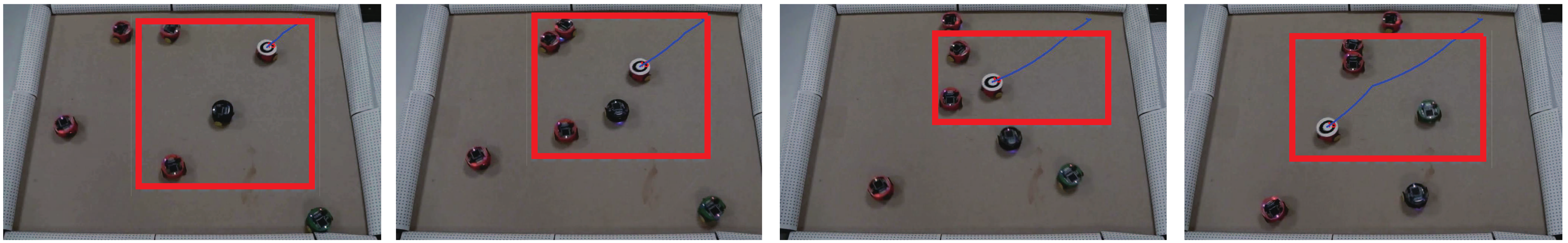}}
		\label{r2r3}}
	\caption{Snapshots of the arena tests captured by a top-down camera to demonstrate the robot-to-robot events. A robot localization algorithm \cite{Colias-localization} tracks the trajectory of robot with a specific pattern on top.}
	\label{arena captures}
\end{figure}

The snapshots shown in Fig. \ref{arena captures} demonstrate some key events in the arena tests. More importantly, since we add in new motion features in the proposed biorobotic approach, we define new criterion to calculate the success rate, as shown in Table \ref{sr-table}. Intuitively, in case of collision avoidance to moving robots, the proposed approach shows much higher success rates than the former models tested at different speeds and density of robots in dynamic robot scenes. It also appears that the proposed model is weaker in collision avoiding to the peripheries of the arena compared to the LGMD2 model. The reason is that some angular approaching to the periphery patterns could highly activate the DSN-R or the DSN-L, so that inhibiting both the LGMDs neurons. Interestingly, another achievement of this biorobotic approach is the generation of robot tracking behaviors by the spiking DSNs in dynamic robot scenes.

\section{CONCLUSIONS}
\label{conclusion}
In this paper, we presented a synthetic neural vision system, as an embedded vision system in autonomous micro-robots, for motion pattern recognition in dynamic scenes, in both a timely and accurate manner. The novelty of the proposed biorobotic approach is to design the integration of four neuron models motivated by insects' visual systems. The LGMDs neurons in the locusts are only perceiving looming objects, whilst the combination of LGMD1 and LGMD2 models can discriminate well between the looming and the recession of dark objects. On the other hand, the DSNs in the flies are only sensitive to translational motion. The perfectly complementary functionality of these neurons significantly advance the modeling of motion perception system with the recognition of more abundant motion features in mobile robots. We also design simple robot motion behaviors for indicating the results of different motion pattern recognition.

We have demonstrated the specific characteristics of each neuron in the open-loop robot tests. The spike frequency or activation of each neuron corresponds to a specific motion pattern. Our arena tests with multi-robots validated the effectiveness of this approach in recognizing different motion patterns, timely in dynamic robot scenes. Moreover, compared with two former studies, we verified the enhanced collision selectivity of this neural system with higher collision-detecting success-rate by extracting new motion features.

In our future work, we will continue incorporating other visual neurons in the synthetic neural system with more motion features extracted to enrich the motion pattern `library'. For example, there are also specific neurons in the fly's visual system, which are only sensitive to small object movements. Moreover, we will test the proposed approach with other mobile robot platforms or vehicles in dynamic and complex scenes. Our goal is to build low-cost, low-power and robust neuromorphic sensors using these bio-inspired visual processing methodologies for motion perception.

\bibliographystyle{IEEEtran}

\bibliography{IEEEqinbing}

\begin{thebibliography}{10}
\providecommand{\url}[1]{#1}
\csname url@samestyle\endcsname
\providecommand{\newblock}{\relax}
\providecommand{\bibinfo}[2]{#2}
\providecommand{\BIBentrySTDinterwordspacing}{\spaceskip=0pt\relax}
\providecommand{\BIBentryALTinterwordstretchfactor}{4}
\providecommand{\BIBentryALTinterwordspacing}{\spaceskip=\fontdimen2\font plus
\BIBentryALTinterwordstretchfactor\fontdimen3\font minus
  \fontdimen4\font\relax}
\providecommand{\BIBforeignlanguage}[2]{{%
\expandafter\ifx\csname l@#1\endcsname\relax
\typeout{** WARNING: IEEEtran.bst: No hyphenation pattern has been}%
\typeout{** loaded for the language `#1'. Using the pattern for}%
\typeout{** the default language instead.}%
\else
\language=\csname l@#1\endcsname
\fi
#2}}
\providecommand{\BIBdecl}{\relax}
\BIBdecl

\bibitem{ECCV-2016}
H.~Kim, S.~Leutenegger, and A.~J. Davison, ``Real-time 3d reconstruction and
  6-dof tracking with an event camera,'' in \emph{ECCV}, 2016, Conference
  Proceedings.

\bibitem{multi-motion-2016}
R.~Sabzevari and D.~Scaramuzza, ``Multi-body motion estimation from monocular
  vehicle-mounted cameras,'' \emph{IEEE Transactions on Robotics}, 2016.

\bibitem{CVPR-2016}
J.~Redmon and S.~Divvala, ``You only look once: Unified, real-time object
  detection,'' in \emph{CVPR}, 2016, Conference Proceedings.

\bibitem{flownet-2015}
A.~Dosovitskiy, P.~Fischer, E.~Ilg, P.~Hausser, C.~Hazirbas, and V.~Golkov,
  ``Flownet: Learning optical flow with convolutional networks,'' in
  \emph{ICCV}, 2015, Conference Proceedings.

\bibitem{NeuroVisionSensor-2000}
G.~Indiveri and R.~Douglas, ``Neuromorphic vision sensors,'' \emph{Science},
  2000.

\bibitem{FlyingInsects-2010}
D.~Floreano, J.-C. Zufferey, M.~V. Srinivasan, and C.~Ellington, \emph{Flying
  Insects and Robots}.\hskip 1em plus 0.5em minus 0.4em\relax Springer, 2010.

\bibitem{Serres-2017}
J.~R. Serres and F.~Ruffier, ``Optic flow-based collision-free strategies: From
  insects to robots,'' \emph{Arthropod Structure \& Development}, 2017.

\bibitem{IROS-LGMDs}
Q.~Fu, C.~Hu, T.~Liu, and S.~Yue, ``Collision selective lgmds neuron models
  research benefits from a vision-based autonomous micro robot,'' in
  \emph{IROS}, 2017, Conference Proceedings.

\bibitem{LGMDs-2016}
F.~C. Rind, S.~Wernitznig, P.~Polt, A.~Zankel, D.~Gutl, J.~Sztarker, and
  G.~Leitinger, ``Two identified looming detectors in the locust: ubiquitous
  lateral connections among their inputs contribute to selective responses to
  looming objects,'' \emph{Scientific Reports}, 2016.

\bibitem{LGMD-car-2017}
M.~Hartbauer, ``Simplified bionic solutions: a simple bio-inspired vehicle
  collision detection system,'' \emph{Bioinspiration and Biomimetics}, 2017.

\bibitem{LGMD1-robot2010}
S.~Yue, R.~D. Santer, Y.~Yamawaki, and F.~C. Rind, ``Reactive direction control
  for a mobile robot: a locust-like control of escape direction emerges when
  a bilateral pair of model locust visual neurons are integrated,''
  \emph{Autonomous Robots}, 2010.

\bibitem{Colias-Hu}
C.~Hu, F.~Arvin, C.~Xiong, and S.~Yue, ``Bio-inspired embedded vision system
  for autonomous micro-robots: The lgmd case,'' \emph{IEEE Transactions on
  Cognitive and Developmental Systems}, 2017.

\bibitem{LGMD2-Fu}
Q.~Fu and S.~Yue, ``Modelling lgmd2 visual neuron system,'' in \emph{2015 IEEE
  25th International Workshop on MLSP}, Conference Proceedings.

\bibitem{LGMD2-BMVC}
Q.~Fu, C.~Hu, and S.~Yue, ``Bio-inspired collision detector with enhanced
  selectivity for ground robotic vision system,'' in \emph{BMVC 2016},
  Conference Proceedings.

\bibitem{LGMD1-Yue2006}
S.~Yue and F.~Claire~Rind, ``Visual motion pattern extraction and fusion for
  collision detection in complex dynamic scenes,'' \emph{Computer Vision and
  Image Understanding}, 2006.

\bibitem{DSN-IJCNN}
Q.~Fu and S.~Yue, ``Modeling direction selective visual neural network with on
  and off pathways for extracting motion cues from cluttered background,'' in
  \emph{The 2017 IJCNN}, 2017, Conference Proceedings.

\bibitem{Circuit-motion}
A.~Borst and T.~Euler, ``Seeing things in motion: models, circuits, and
  mechanisms,'' \emph{Neuron}, 2011.

\bibitem{Borst-common}
A.~Borst and M.~Helmstaedter, ``Common circuit design in fly and mammalian
  motion vision,'' \emph{nature neuroscience}, 2015.

\bibitem{EMD-1989}
A.~Borst and M.~Egelhaaf, ``Principles of visual motion detection,''
  \emph{Trends Neurosci}, 1989.

\bibitem{LGMD1-DSN-competing}
S.~Yue and F.~C. Rind, ``Redundant neural vision systems—competing for
  collision recognition roles,'' \emph{IEEE Transactions on Autonomous Mental
  Development}, 2013.

\bibitem{DSN-2013}
------, ``Postsynaptic organization of directional selective visual neural
  networks for collision detection,'' \emph{Neurocomput}, 2013.

\bibitem{LGMD-DSNs-Collision}
G.~Zhang, C.~Zhang, and S.~Yue, ``Lgmd and dsns neural networks integration for
  collision predication,'' in \emph{The 2016 IJCNN}, 2016, Conference
  Proceedings, pp. 1174--1179.

\bibitem{ROBIO-2017}
Q.~Fu and S.~Yue, ``Mimicking fly motion tracking and fixation behaviors with a
  hybrid visual neural network,'' in \emph{IEEE Int. Conf. on Robotics and
  Biomimetics}, 2017, Conference Proceedings.

\bibitem{Colias-localization}
T.~Krajn{\' i}k, M.~Nitsche, I.~Faigl, P.~Van{\v e}k, M.~Saska, L.~P{\v r}eu{\v
  c}il, T.~Duckett, and M.~Marta, ``A practical multirobot localization
  system,'' \emph{Journal of Intelligent \& Robotic Systems}, 2014.

\end{thebibliography}
\addtolength{\textheight}{-12cm}  
\end{document}